\documentclass[utf8]{frontiersSCNS} % for Science, Engineering and Humanities and Social Sciences articles
\usepackage{url,hyperref,lineno,microtype,subcaption}
\usepackage[onehalfspacing]{setspace}
\usepackage{soul}

\def\keyFont{\fontsize{8}{11}\helveticabold }
\def\firstAuthorLast{Dunn} %use et al only if is more than 1 author
\def\Authors{Jonathan Dunn\,$^{1,*}$}
\def\Address{$^{1}$Department of Linguistics, University of Canterbury, Christchurch, New Zealand 
 }
\def\corrAuthor{Corresponding Author}
\def\corrEmail{jonathan.dunn@canterbury.ac.nz}

\begin{document}
\onecolumn
\firstpage{1}

\title[Global Syntactic Variation]{Global Syntactic Variation in Seven Languages: Towards a Computational Dialectology} 

\author[\firstAuthorLast ]{\Authors} %This field will be automatically populated
\address{} %This field will be automatically populated
\correspondance{} %This field will be automatically populated

\extraAuth{}% If there are more than 1 corresponding author, comment this line and uncomment the next one.
%\extraAuth{corresponding Author2 \\ Laboratory X2, Institute X2, Department X2, Organization X2, Street X2, City X2 , State XX2 (only USA, Canada and Australia), Zip Code2, X2 Country X2, email2@uni2.edu}

\maketitle

\begin{abstract}
\section{}

The goal of this paper is to provide a complete representation of regional linguistic variation on a global scale. To this end, the paper focuses on removing three constraints that have previously limited work within dialectology/dialectometry. First, rather than assuming a fixed and incomplete set of variants, we use Computational Construction Grammar to provide a replicable and falsifiable set of syntactic features. Second, rather than assuming a specific area of interest, we use global language mapping based on web-crawled and social media datasets to determine the selection of national varieties. Third, rather than looking at a single language in isolation, we model seven major languages together using the same methods: Arabic, English, French, German, Portuguese, Russian, and Spanish. Results show that models for each language are able to robustly predict the region-of-origin of held-out samples better using Construction Grammars than using simpler syntactic features. These global-scale experiments are used to argue that new methods in \textit{computational sociolinguistics} are able to provide more generalized models of regional variation that are essential for understanding language variation and change at scale.

\tiny
 \keyFont{ \section{Keywords:} dialectology, dialectometry, construction grammar, syntactic variation, text classification, language mapping, dialect mapping, computational sociolinguistics} %All article types: you may provide up to 8 keywords; at least 5 are mandatory.
\end{abstract}

\section{Introduction}

This paper shows that computational models of syntactic variation provide precise and robust representations of national varieties that overcome the limitations of traditional survey-based methods. A computational approach to variation allows us to systematically approach three important problems: First, what set of variants do we consider? Second, what set of national dialects or varieties do we consider? Third, what set of languages do we consider? These three questions are usually answered in reference to the \textit{convenience} or \textit{interests} of the research project at hand. From that perspective, the goal of this paper is global, multi-lingual, whole-grammar syntactic dialectometry. Previous work has performed whole-grammar dialectometry with Construction Grammars, first using a pre-defined inventory of national varieties \citep{d18b} and then using data-driven language mapping to select the inventory of national varieties \citep{Dunn2019a}. This paper further extends computational dialectometry by studying seven languages across both web-crawled and social media corpora. The paper shows that a classification-based approach to syntactic variation produces models that (i) are able to make accurate predictions about the region-of-origin of held-out samples, (ii) are able to characterize the aggregate syntactic similarity between varieties, and (iii) are able to measure the uniqueness of varieties as an empirical correlate for qualitative notions like \textit{inner-circle} vs. \textit{outer-circle}.

What features do we use for dialectometry? Most previous work relies on phonetic or phonological features \citep{k92, k96, h04, lab05, n06, n09, gsg11, wn11, g13, nk13, JackGrieve2013, kjb14, wn15, kv18} for the simple reason that phonetic representations are relatively straight-forward: a vowel is a vowel and the measurements are the same across varieties and languages. Previous work on syntactic variation has focused on either (i) an incomplete set of language-specific variants, ranging from only a few features to hundreds \citep{s07, s09, s10, Grieve2011, c12, Grieve2012, s13, s14, g16, ss16, sghr16, cr17, gs18, t18} or (ii) language-independent representations such as function words \citep{ak13} or sequences of part-of-speech labels \citep{hf07,k}. This forces a choice between either an \textit{ad hoc} and incomplete syntactic representation or a reproducible but indirect syntactic representation.

This previous work on syntactic dialectometry has depended on the idea that a grammar is an inventory of specific structures: the double-object construction versus the prepositional dative, for example. Under this view, there is no language-independent feature set for syntax in the way that there is for phonetics. But we can also view syntax from the perspective of a discovery-device grammar \citep{Chomsky1957, Goldsmith2015}: in this case, our theory of grammar is not a specific description of a language like English but rather a function for mapping between observations of English and a lower-level grammatical description of English: $G=D({\textsc{corpus}})$. Thus, a discovery-device grammar ($G$) is an abstraction that represents what the grammatical description would be if we applied the learner ($D$) to a specific sample of the language ($\textsc{corpus}$). A discovery-device grammar allows us to generalize syntactic dialectometry: we are looking for a model of syntactic variation, $V$, such that when applied to a grammar, $V(G)$, the model is able to predict regional variation in the grammar. But $G$ is different for each language, so we generalize this to $V(D(\textsc{corpus}))$. In other words, we use an independent corpus for each language as input to a discovery-device grammar and then use the resulting grammar as a feature space for performing dialectometry. This approach, then, produces an inventory of syntactic features for each language in a reproducible manner in order to replace hand-crafted syntactic features. The specifics of the datasets used for modeling regional variation are described in Section 2.1 and the discovery-device grammar used to create reproducible feature sets is described in Section 2.2.

What type of model should we use to represent global syntactic variation? Previous work has relied largely on unsupervised methods like clustering \citep{wn11}, factor analysis of spatial autocorrelation scores \citep{g13}, and individual differences scaling as an extension of multidimensional scaling \citep{Ruette2014}. These models attempt to aggregate individual variants into larger bundles of features: which individual features represent robust aggregate isoglosses with a similar geographic extent? The problem is that it is difficult to evaluate the predictions of one such bundle against another. While useful for visualizations, these models are difficult to evaluate against ground-truths. Another strand of work models the importance of predictor variables on the use of a particular variant, with geographic region as one possible predictor \citep{sghr16}. These models are based on multivariate work in sociolinguistics that attempts to find which linguistic, social, or geographic features are most predictive of a particular variant. 

While useful for understanding individual variants, however, these models are unable to handle the aggregation of variants directly. For example, although it is possible to create a distance matrix between regions for each individual feature and then to aggregate these matrices, the resulting aggregations are subject to variability: What is the best aggregation method? If two methods provide different maps, which should we prefer? How stable are aggregations across folds? On the one hand, we want dialectometry to establish a ground-truth about the regional distribution of variants and dialects. But, on the other hand, because unsupervised methods like clustering are subject to such potential variability, we also need a ground-truth to evaluate which aggregation method is the most accurate.

One solution to this problem is to take a classification approach, in which the ground-truth is the region-of-origin for individual samples. Given a model of dialectal variation, how accurately can that model predict the region-of-origin of new samples? For example, the idea is that a more complete description of the syntactic differences between Australian English and New Zealand English will be able to predict more accurately whether a new sample comes from Australia or New Zealand. This prediction task provides a ground-truth for aggregation. But it comes with two important caveats: First, a high prediction accuracy does not guarantee that the model captures all relevant variation, only that it captures enough variation to distinguish between national varieties. This can be mitigated, however, by using cross-fold validation and unmasking as shown in Section 3.2. Second, while most work in dialectometry tries to establish geographic boundaries, this work assumes geographic boundaries (i.e., polygons of nation-states).

What languages and regions need to be represented in dialectometry? Because of coloniziation and globalization \citep{k90}, a few languages like English are now used around the world by diverse national communities. Even though these international languages have global speech communities, dialectology and sociolinguistics continue to focus largely on sub-national dialects, often within so-called \textit{inner-circle} varieties \citep{Kachru1982}. This paper joins recent work in taking a global approach by using geo-referenced texts \citep{geq12, df15, ds17} to represent national varieties \citep{sghr16, cr17, Cook2017, Rangel2017, t18, d18b, Dunn2019a}. The basic point is that in order to represent regional variation as a complete system, dialectometry must take a global perspective. This paper uses data-driven language mapping to choose (i) which international languages are used widely enough to justify inclusion and (ii) which languages in which countries need to be included as national varieties. We use geo-referenced corpora drawn from web pages and social media for both tasks. Seven languages are selected for dialectometry experiments: Arabic, English, French, German, Portuguese, Russian, and Spanish. These seven languages account for 59.25\% of the web-crawled corpus and 74.67\% of the social media corpus. The corpora are regionalized to countries. Thus, the assumption is that any country which frequently produces data in a language has a national variety of that language. For example, whether or not there is a distinct variety of New Zealand English depends entirely on how much English data is observed from New Zealand in these datasets. The models then have the task of determining how distinct New Zealand English is from other national varieties of English.

First, we consider the selection of (i) languages and (ii) national varieties of languages (Section 2.1) as well as the selection of a syntactic feature space (Section 2.2). We then present the specifics of the experimental framework (Section 2.3). Second, we compare prediction accuracies by language and feature set (Section 3.1), in order to measure the quality of the models. Next, we evaluate the robustness of the models across rounds of feature pruning and the similarity of the models across registers in order to examine potential confounds (Section 3.2). Having validated the models themselves, the next section examines regional accuracies and the similarities between national varieties (Section 3.3). Finally, we develop measures for the syntactic uniqueness of each regional variety (Section 3.4) and search for empirical correlates of concepts like \textit{inner-circle} and \textit{outer-circle} within this corpus-based approach (Section 3.5). Third, we discuss two important issues: the application of different categorizations like \textit{inner-circle vs. outer-circle} or \textit{native vs. non-native} to these datasets (Section 4.1) and the implications of a computational approach to dialectometry for sociolinguistics more broadly (Section 4.2).

\section{Materials and Methods}

\subsection{Language Mapping and Dialectometry}

We begin with data-driven language mapping: First, what languages have enough national varieties to justify modeling? Second, which national varieties should be included for each language? Third, which datasets can be used to represent specific national varieties and how well do these datasets represent the underlying populations? This paper depends on geo-referenced corpora: text datasets with meta-data that ties each document to a specific place. The size of both datasets by region is shown in Table 1, together with ground-truth population data from the UN \citep{UnitedNations2017}. The size of each region relative to the entire dataset is also shown: for example, 14.6\% of the web corpus comes from Western Europe which accounts for only 5.7\% of the global population. This comparison reveals the over-representation and under-representation of each region.

\begin{table}
\begin{center}
	\def\arraystretch{1.5}
	\begin{tabular}{|l|c|rr|rr|rr|}
		\hline
		\textbf{Region} & \textbf{Countries} & \textbf{Population} & \textbf{(\%)} & \textbf{Web} & \textbf{(\%)} & \textbf{Twitter} & \textbf{(\%)} \\
		\hline
		Africa, North & 9 & 250 mil & 3.4\% & 123.85 mil & 0.7\% & 85.55 mil & 2.1\% \\
		Africa, Southern & 4 & 75 mil & 1.0\% & 59.07 mil & 0.4\% & 87.34 mil & 2.1\% \\
		Africa, Sub-Saharan & 73 & 742 mil & 10.1\% & 424.75 mil & 2.6\% & 254.20 mil & 6.1\% \\
		\hline
		America, Brazil & 1 & 206 mil & 2.8\% & 218.11 mil & 1.3\% & 118.13 mil & 2.9\% \\
		America, Central & 25 & 214 mil & 2.9\% & 886.61 mil & 5.3\% & 383.81 mil & 9.3\% \\
		America, North & 2 & 355 mil & 4.8\% & 236.59 mil & 1.4\% & 350.12 mil & 8.5\% \\
		America, South & 11 & 210 mil & 2.9\% & 1,163.00 mil & 7.0\% & 402.15 mil & 9.7\% \\
		\hline
		Asia, Central & 10 & 198 mil & 2.7\% & 965.09 mil & 5.8\% & 102.79 mil & 2.5\% \\
		Asia, East & 8 & 1,635 mil & 22.3\% & 2,201.86 mil & 13.2\% & 95.70 mil & 2.3\% \\
		Asia, South & 7 & 1,709 mil & 23.3\% & 448.23 mil & 2.7\% & 331.19 mil & 8.0\% \\
		Asia, Southeast & 22 & 615 mil & 8.4\% & 2,011.06 mil & 12.1\% & 245.18 mil & 5.9\% \\
		\hline
		Europe, East & 17 & 176 mil & 2.4\% & 4,553.10 mil & 27.4\% & 322.46 mil & 7.8\% \\
		Europe, Russia & 1 & 144 mil & 2.0\% & 101.44 mil & 0.6\% & 105.04 mil & 2.5\% \\
		Europe, West & 25 & 421 mil & 5.7\% & 2,422.85 mil & 14.6\% & 823.80 mil & 19.9\% \\
		\hline
		Middle East & 15 & 334 mil & 4.5\% & 660.73 mil & 4.0\% & 222.98 mil & 5.4\% \\
		\hline
		Oceania & 8 & 59 mil & 1.0\% & 164.02 mil & 1.0\% & 213.06 mil & 5.1\% \\
		\hline
		\textsc{\textbf{total}} & \textbf{199} & \textbf{7.35 bil} & \textbf{100\%} & \textbf{16.65 bil} & \textbf{100\%} & \textbf{4.14 bil} & \textbf{100\%}\\
		\hline
	\end{tabular}
	\caption{Size of Geo-Referenced Corpora in Words by Region}
	\label{tab:1}
	\end{center}
\end{table}

Data comes from two sources of digital texts: web pages from the Common Crawl\footnote{\url{http://www.commoncrawl.org}} and social media from Twitter\footnote{\url{http://www.twitter.com}}. The Common Crawl data represents a large snapshot of the internet; although we cannot direct the crawling procedures, we are able to process the archived web pages from the perspective of a geo-referenced corpus. The author of each individual web page may be unknowable but we can use country-specific top-level domains for country-level geo-referencing: for example, web pages under the \textit{.nz} domain are from New Zealand. Previous work has shown that there is a relationship between domain-level geo-referenced web pages and national varieties \citep{Cook2017}. Some countries are not available because their top-level domains are used for other purposes (i.e., \textit{.ai}, \textit{.fm}, \textit{.io}, \textit{.ly}, \textit{.ag}, \textit{.tv}). Domains that do not contain geographic information are also removed from consideration (e.g., \textit{.com} sites). The Common Crawl dataset covers 2014 through the end of 2017, totalling 81.5 billion web pages. As shown in Table 1, after processing this produces a corpus of 16.65 billion words. This dataset represents 166 out of 199 total countries considered in this paper. Some countries do not use their country-level domains as extensively as others: in other words, \textit{.us} does not account for the same proportion of web pages from the United States as \textit{.nz} does from New Zealand. It is possible that this skews the representation of particular areas. Thus, Table 1 shows the UN-estimated population for each region as reference. The web corpus is available for download\footnote{\url{https://labbcat.canterbury.ac.nz/download/?jonathandunn/CGLU\_v3}} as is the code used to create the corpus.\footnote{\url{https://github.com/jonathandunn/common_crawl_corpus}}

\begin{figure*}
\begin{center}
	\fbox{
	\includegraphics[scale=0.365]{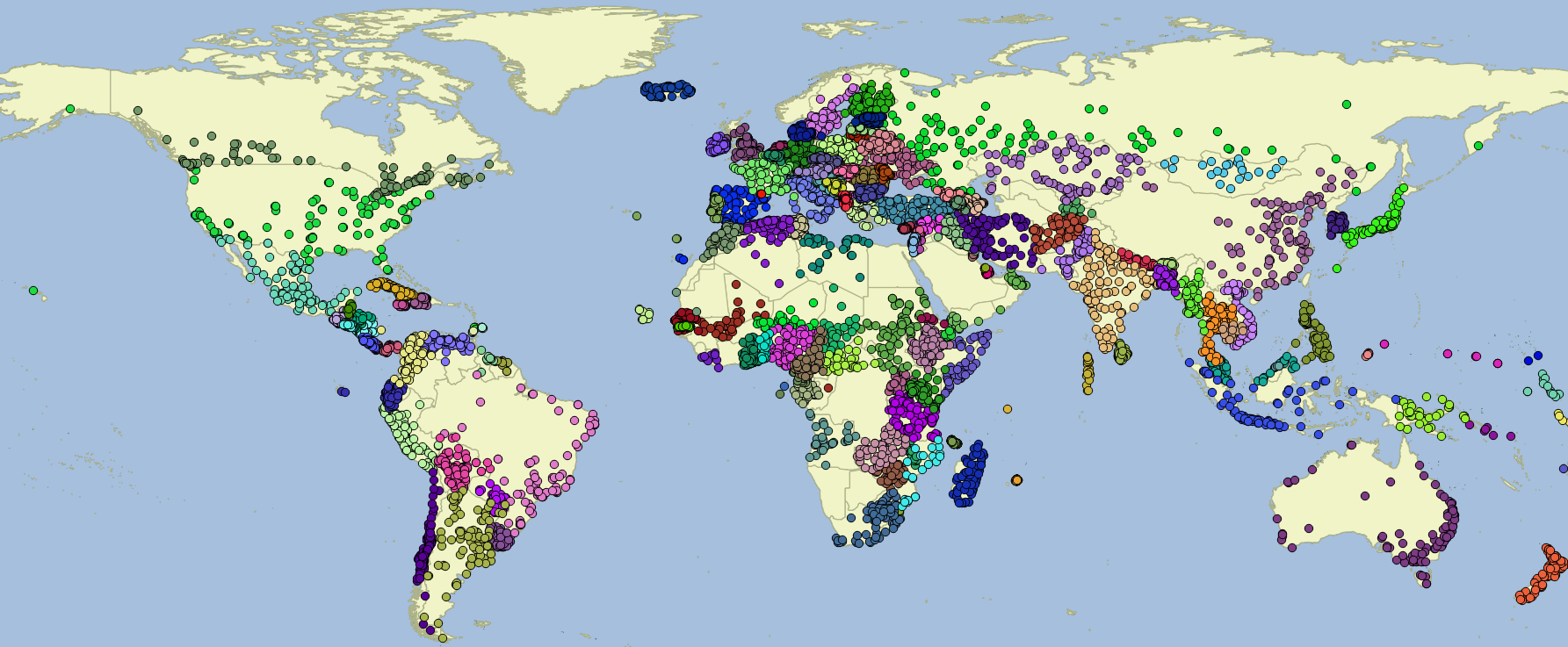}
	}
	\caption{Cities for Collection of Twitter Data (50km radius from each)}
\end{center}
\end{figure*}

In isolation, web-crawled data provides one observation of global language use. Another common source of data used for this purpose is Twitter (e.g., Eisenstein, et al., 2010; Roller, et al., 2012; Kondor, et al., 2013; Mocanu, et al., 2013; Eisenstein, et al., 2014; Graham, et al., 2014; Donoso \& Sanchez, 2017). The shared task at PAN-17, for example, used Twitter data to represent national varieties of several languages \citep{Rangel2017}. A spatial search is used to collect Tweets from within a 50km radius of 10k cities.\footnote{\url{https://github.com/datasets/world-cities}} This city-based search avoids biasing the selection by using language-specific keywords or hashtags. A map of each city used for collection is shown in Figure 1; while this approach avoids a language-bias, it could under-represent rural areas given the 50 km radius of each collection area. The Twitter data covers the period from May of 2017 until early 2019, drawn from the Twitter API using a spatial query. This creates a corpus containing 1,066,038,000 Tweets. The language identification component, however, only provides reliable predictions for samples containing at least 50 characters (c.f., the language id code\footnote{\url{https://github.com/jonathandunn/idNet}} and the models used\footnote{\url{https://labbcat.canterbury.ac.nz/download/?jonathandunn/idNet_models}}). Thus, the corpus is pruned to include only Tweets above that length threshold. As shown in Table 1, this produces a corpus containing 4.14 billion words. While the Common Crawl corpus represents 166 countries, the Twitter corpus represents 169. There are 33 countries that only Twitter represents (not the Common Crawl) and 30 that only the Common Crawl represents (not Twitter). This shows the importance of drawing on two different sources of language use.

Given the idiosyncracies of these two datasets (i.e., the availability of country-codes for web data and the selection of cities for Twitter data), it is quite likely that each represents different populations or, at least, that each represents different registers of language usage from the same population. We can use ground-truth population data to deal with the problem of different populations. First, notice that both datasets under-represent all regions in Africa; but the web dataset has the worst under-representation: while Africa accounts for 14.5\% of the world's population, it accounts for only 3.7\% of the web corpus. The Americas and Europe, on the other hand, are over-represented in both datasets. Twitter especially over-represents North America (8.5\% of the corpus vs. 4.8\% of the population); but the web corpus under-represents North America (only 1.4\% of the corpus), mostly from the lack of adoption of the \textit{.us} domain. Western Europe is over-represented in both corpora: while it acounts for only 5.7\% of the population, it provides 14.6\% of the web corpus and 19.9\% of the Twitter corpus. Although these trends are expected, it is helpful to quantify the degree of over-representation. Less expectedly, the web corpus greatly over-represents Eastern Europe (27.4\% of the corpus but only 2.4\% of the population). Asia, especially the East and South, are under-represented in both datasets.

On the one hand, the use of population data here allows us to quantify exactly how each of these datasets is skewed. On the other hand, our purpose is to model regional syntactic variation: do the datasets need to be prefectly aligned with regional populations in order to achieve this? There are two observations to be made: First, if a region is over-represented then we do not need to worry about missing any national varieties from that area; but we should be worried about over-representing those particular national varieties (this is why there is a cap on the number of training samples from each dialect). Second, it could be the case that we are missing national varieties from under-represented areas. For example, any missing national varieties are likely to be from Africa or East Asia, given the skewed representations of this dataset. Related work, however, has shown that it in the case of major international languages like those considered here, the problem is over-representation rather than under-representation in the form of missing regional varieties \citep{Dunn2019c}. We leave it to future work to make improvements in the selection of regional varieties using population-based sampling to overcome skewness in corpus distributions.

What languages should be included in a model of global syntactic variation? Given that we are using countries to define regional varieties, a language needs to occur in many countries. Here we use a threshold of 1 million words to say that a language is used significantly in a given country. Table 2 shows the seven languages included in this study, encompassing 59.25\% of the web corpus and 74.67\% of the Twitter corpus. Some other languages occur in several countries in one dataset but not the other and so are not included. For example, Italian occurs in 17 countries in the web corpus but only 2 in the Twitter corpus; Indonesian occurs in 10 countries in the web corpus but only 3 countries in the Twitter corpus. Given that we model varieties using a classifier, we focus on those languages that have a sufficient number of national varieties to make classification a meaningful approach.

\subsection{Finding Syntactic Variants}

This paper represents syntactic variants using a discovery-device Construction Grammar (CxG) that produces a CxG for each language given an independent corpus representing that language. CxG itself is a usage-based paradigm that views grammar as a set of overlapping constructions made up of slot-fillers defined by syntactic, semantic, and sometimes lexical constraints \citep{g06a, l08}. This paper draws on recent approaches to computational modeling of CxGs \citep{d17, d18, Dunn2019}, including previous applications of a discovery-device CxG to dialectometry for English \citep{d18b, Dunn2019a}.

Constructions are represented as a sequence of slot-constraints, as in (1a). Slots are separated by dashes and constraints are defined by both type (Syntactic, Joint Semantic-Syntactic, Lexical) and by filler (for example: \textsc{noun}, a part-of-speech or \textsc{animate}, a semantic domain).

\noindent(1a) [SYN:\textsc{noun} --- SEM-SYN:\textsc{transfer[V]} --- SEM-SYN:\textsc{animate[N]} --- SYN:\textsc{noun}] \\
(1b) ``He gave Bill coffee." \\
(1c) ``He gave Bill trouble." \\
(1d) ``Bill sent him letters." \\
(2a) [SYN:\textsc{noun} --- LEX:``give" --- SEM-SYN:\textsc{animate[N]} --- LEX:``a hand"] \\
(2b) ``Bill gave me a hand." \\

The construction in (1a) contains four slots: two with joint semantic-syntactic constraints and two with simple syntactic constraints. The examples in (1b) to (1d) are tokens of the construction in (1a). Lexical constraints, as in (2a), represent idiomatic sentences like (2b). A CxG is a collection of many individual constructions. For the purposes of dialectometry, these are quantified as one-hot encodings of construction frequencies. This, in essence, provides a bag-of-constructions that is evaluated against traditional bag-of-words features.

A large portion of the language-learning corpus for each language comes from web-crawled data \citep{bbfz09, m12, b14} and data from the CoNLL 2017 Shared Task \citep{ghl17}. Because the goal is to provide a wide representation of each language, this is augmented by legislative texts from the EU and UN \citep{Tiedemann2012, Skadins2014}, the OpenSubtitles corpus \citep{Tiedemann2012}, and newspaper texts. The exact collection of documents used for learning CxGs is available for download.\footnote{\url{https://labbcat.canterbury.ac.nz/download/?jonathandunn/CxG\_Data\_FixedSize}} While both web-crawled and social media datasets are used to represent national varieties, the grammars used are learned mainly from web-crawled corpora. On the one hand, we use separate datasets for grammar learning and dialectometry in order to remove the possible confound that the grammars are over-fitting a specific dataset. On the other hand, we do not explicitly know which regional varieties the data used for grammar learning is drawn from. The discussion in Section 3.5, as well as other work \citep{Dunn2019a}, shows that at least the English grammar better represents inner-circle varieties like UK English. In this case, then, we prefer to avoid the possible confound of over-fitting even though the result is a grammar that is learned from datasets implicitly drawn from inner-circle varieties.

This paper evaluates two alternate CxGs for dialectometry, alongside function words and lexical features: CxG-1 \citep{d18, d18b} and CxG-2 \citep{Dunn2019}. As described and evaluated elsewhere \citep{Dunn2019}, CxG-1 relies on frequency to select candidate slot-constraints while CxG-2 relies on an association-based search algorithm. The differences between the two competing discovery-device grammars as implementations of different theories of language learning are not relevant here. Rather, we evaluate both grammars because previous work \citep{d18b} relied on CxG-1 and this comparison makes it possible to connect the multi-lingual experiments in this paper with English-only experiments in previous work. It should be noted, however, that other work has shown that association-based constraints out-perform frequency-based constraints across several languages \citep{Dunn2019}. As shown in Section 3, this paper finds that association-based constraints also perform better on the task of dialectometry. This is important because the evaluation connects the emergence of syntactic structure with variation in syntactic structure.

Previous work on syntactic dialectometry focuses on paired sets of features which can be viewed as alternate choices that express the same function or meaning. In other words, these approaches contrast constructions like the double object vs. the prepositional dative  and then quantify the relative preference of particular varieties for one variant over the other. From our perspective, such an approach is essential for a limited feature space because syntactic variation is structured around different constructions that encode the same function or meaning. In other words, two constructions which have entirely different uses cannot be in competition with one another: constrasting the double object and the get-passive constructions, in isolation, is not a meaningful approach to syntactic variation because their frequencies are influenced by other unseen parts of the grammar. On the other hand, looking at the frequency of a single construction in isolation can be meaningful but will never reveal the full picture of syntactic variation.

This whole-grammar construction-based approach to dialectology represents as much of the functional space as possible. This provides an implicit pairing of syntactic variants: without a topic bias, we expect that the relative frequency of a specific construction will be consistent across documents. If one construction is more frequent, that indicates an increased preference for that construction. This approach does not explicitly pair variants because part of the problem is to learn which constructions are in alternation. From a different perspective, we could view alternating variants as knowledge that is traditionally given to models within quantitative sociolinguistics: which constructions are in competition with one another? But the idea here is to leave it to the model itself to determine which constructions are in competition.

Because this work is situated within both dialectometry and construction grammar, we view syntactic variation as fundamentally structured around function and meaning (as described above). But more traditional sociolinguistic and generativist work on syntactic variation does not share this underlying view. In this case the prediction task itself allows us to translate between competing assumptions: regardless of how we understand the source of variation, the models are ultimately evaluated on how well they are able to predict region-of-origin (samples from New Zealand vs. samples from Australia) using only syntactic representations. This type of ground-truth evaluation can be undertaken, with greater or lesser success, with any set of assumptions. Whether or not dialectal variation is fundamentally based on alternations and whether or not dialectometry models require alternations, the argument here is that the ability to distinguish between dialects (without topic-based features) is a rigorous evaluation of the quality of a model of dialects.

Finally, how does geographic variation as modeled here interact with register variation? We can think about this in two different ways: First, does register variation within these datasets present a confound by being structured geographically? In other words, if the corpus from Australia represents newspaper and magazine articles but the corpus from New Zealand represents discussion forums, then the ability to distinguish between the two is a confound. Given the size of the datasets, the consistent collection methodology, the cross-fold validation experiments, the large number of national varieties per language, and the comparison of web-based and Twitter data, however, this confound is not likely. Second, is register variation the same underlying phenomenon as regional variation? In other words, is the difference between New Zealand English and Australian English ultimately the same type of phenomenon as the structured difference between newspaper writing and discussion forums? This is an empirical question for future work that requires a dataset containing both register meta-data and spatial meta-data.

\subsection{Modeling National Varieties}

The experiments in this paper take a classification approach to dialectometry: given a one-hot encoding of construction frequencies (i.e., a bag-of-constructions), can we distinguish between different national varieties of a language? There are two main advantages to this approach: First, the model can be evaluated using prediction accuracies on held-out testing data. This is important to ensure that the final model is meaningful. Second, a classification approach provides an implicit measure of the degree of syntactic separation between national varieties across the entire grammar (c.f., region similarities in Section 3.3). A particular construction may be unique to a given variety, but this in itself is less meaningful if the varieties are otherwise the same. How deep or robust is the syntactic variation? How distinct are the national varieties? Dialectometry is about going beyond variation in individual syntactic features to measure the aggregate syntactic relationships between varieties.

\begin{table}
\begin{center}
\def\arraystretch{1.5}
	\begin{tabular}{|l|cr|cr|}
		\hline
		\textbf{Language} & \textbf{Countries (Web)} & \textbf{Words (Web)} & \textbf{Countries (Twitter)} & \textbf{Words (Twitter)} \\
		\hline
		Arabic (ara) & 19 & 348,671,000 & 25 & 179,473,000 \\
		English (eng) & 130 & 4,990,519,000 & 137 & 1,552,268,000 \\
		French (fra) & 36 & 479,857,000 & 24 & 176,009,000 \\
		German (deu) & 24 & 500,029,000 & 7 & 71,234,000 \\
		Portuguese (por) & 14 & 431,884,000 & 22 & 199,080,000 \\
		Russian (rus) & 37 & 1,361,331,000 & 9 & 126,834,000 \\
		Spanish (spa) & 43 & 1,757,200,000 & 44 & 789,239,000 \\
		\hline
		~ & \textit{\% of Total:} & \textit{59.25\%} & \textit{\% of Total:} & \textit{74.67\%} \\
		\hline
		\textbf{Language} & \textbf{Varieties (Web)} & \textbf{N. Test (Web)} & \textbf{Varieties (Twitter)} & \textbf{N. Test (Twitter)} \\
		\hline
		Arabic (ara) & 4 & 14,685 & 7 & 15,537 \\
		English (eng) & 14 & 66,476 & 14 & 64,208 \\
		French (fra) & 13 & 46,562 & 4 & 12,130 \\
		German (deu) & 7 & 35,240 & 2 & 7,722 \\
		Portuguese (por) & 4 & 15,129 & 2 & 8,650 \\
		Russian (rus) & 19 & 84,925 & 3 & 9,164 \\
		Spanish (spa) & 17 & 84,093 & 17 & 76,653 \\
		\hline
	\end{tabular}
	\caption{Above: Number of Countries and Words by Language and Domain \\
	Below: Number of Varieties and Test Samples by Language and Domain}
	\label{tab:1}
	\end{center}
\end{table}

The main set of experiments uses a Linear Support Vector Machine \citep{j98} to classify varieties using CxG features. Parameters are tuned using separate development data.\footnote{Development data allows experimental settings and parameters to be evaluated without over-fitting the training/testing data that is used for the main experiment.} Given the general robust performance of SVMs in the literature relative to other similar classifiers on latent variation tasks \citep{dark16}, we forego a systematic evaluation of classifiers. For reproducibility against future work, all results are calculated on pre-specified training and testing sets. Given the large number of samples in each test set (Table 2) and the robustness in the cross-validation evaluation (Table 4) we are not concerned with over-fitting and given the high performance in general we are not concerned with under-fitting (Table 3). Under this evaluation regime, any classifier could be used; thus, it is not important to contrast a Linear SVM with other shallow classifiers such as Naive Bayes or Decision Trees in this context.  The Linear SVM uses the training data to learn weights for each construction in the grammar for each regional variety; in the aggregate, the model builds a high-dimensional representation of each variety that maximizes the distance between them (i.e., so that varieties like American English and Nigerian English can be easily separated). The quality and generalizability of the models are evaluated using held-out testing data: how well can those same feature weights be used to predict which regional variety a new sample belongs to? Because it is possible here that the varieties could be distinguished in a low-dimensional space (i.e., being separated along only a few constructions), we use unmasking to evaluate the robustness of the models in Section 3.2. This classification-based approach deals very well with the aggregation of features, including being able to ignore redundant or correlated features. On the other hand, this robust aggregation of syntactic features requires that we assume the spatial boundaries of each regional variety.

Moving to data preparation, the assumption is that a language sample from a web-site under the \textit{.ca} domain originated from Canada. This approach to regionalization does not assume that whoever produced that language sample was born in Canada or represents a traditional Canadian dialect group; rather, the assumption is only that the sample represents someone in Canada who is producing language data; but the two are closely related \citep{Cook2017}. This corresponds with the assumption that Twitter posts geo-referenced to particular coordinates represent language use in that place but do not necessarily represent language use by locals. Geo-referenced documents represent language use \textit{in} a particular place. Unlike traditional dialect surveys, however, there is no assurance that individual authors are native speakers \textit{from} that place. We have to assume that most language samples from a given country represent the native varieties of that country. For example, many non-local residents live in Australia; we only have to assume that most speakers observed in Australia are locals. On the one hand, this reflects the difference between corpus-based and survey-based research: we know less about the individuals who are represented in these datasets. On the other hand, this reflects increased mobility: the idea that a \textit{local} individual is born, is raised, and finally dies all in the same location is no longer proto-typical.

In order to average out the influence of out-of-place samples, we use random aggregation to create samples of exactly 1,000 words in both corpora. For example, in the Twitter corpus this means that an average of 59 individual Tweets from a place are combined into a single sample. First, this has the effect of providing more constructions per sample, making the modelling task more approachable. Second and more importantly, individual out-of-place Tweets and web pages are reduced in importance because they are aggregated with other Tweets and web pages presumably produced by local speakers. If we think of non-locals as outliers, this approach aggregates outliers with non-outliers in order to reduce their influence. We leave for future work an evaluation of different approaches to this problem. The larger issue is the relationship between small but carefully curated corpora for which significant meta-data is available for each speaker and these large but noisy corpora which are known to contain out-of-place samples (i.e., tourists in Twitter data). One promising approach is to evaluate such noisy corpora based on how well they are able to predict demographic meta-data for the places they are intended to represent \citep{Dunn2019c}. In this case, it has been shown that web-crawled and Twitter corpora are significantly correlated with population density (especially when controlling for GDP and general rates of internet usage) and that both datasets can be used to predict which languages are used in a country (as represented using census data). While there is much work to be done on this problem, the prediction of demographic meta-data provides a way to evaluate the degree to which large and noisy corpora reflect actual populations.

\begin{figure*}
\begin{center}
\fbox{
	\includegraphics[scale=0.45]{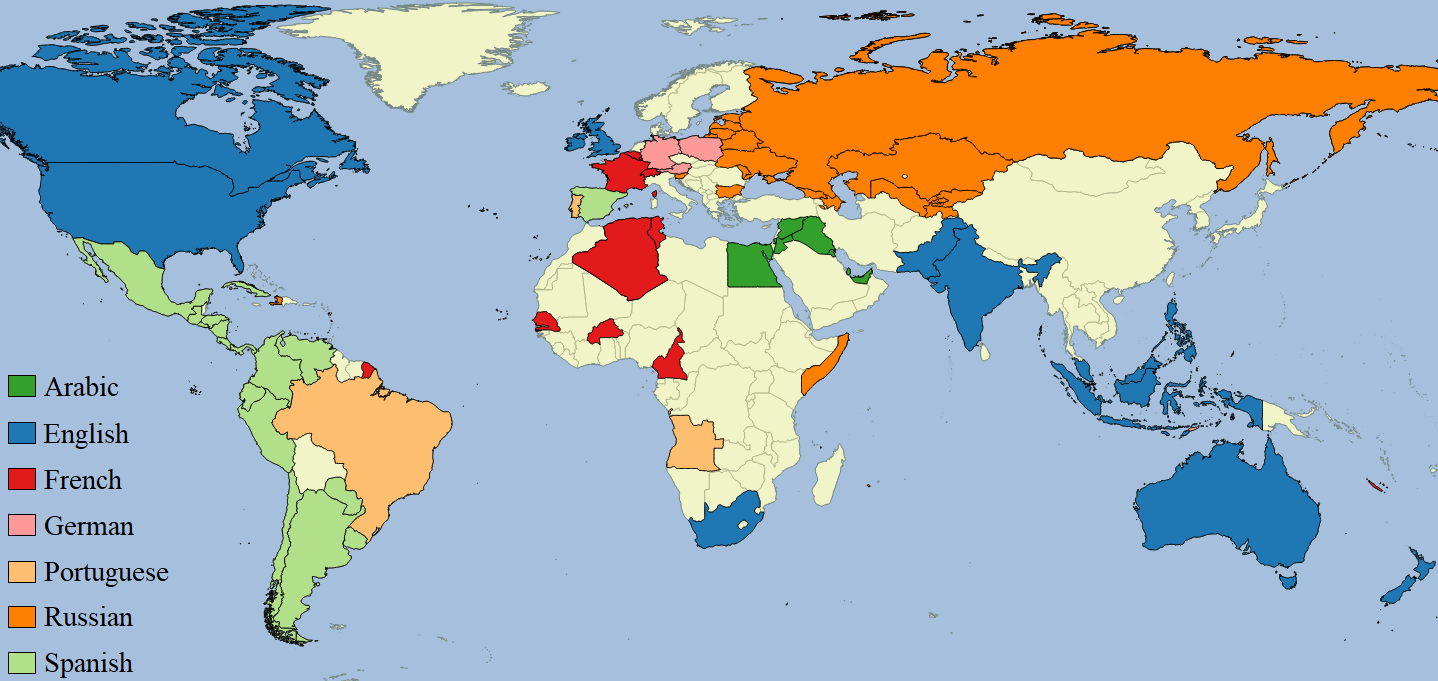}
	}
	\caption{Countries with National Varieties for Selected Languages}
\end{center}
\end{figure*}

We take a simple threshold-based approach to the problem of selecting national varieties to include. For English and Spanish, any national variety that has at least 15 million words in both the Common Crawl and Twitter datasets is included. Given the large number of countries in Table 2, this higher threshold accounts for the fact that both English and Spanish are widely used in these datasets. Lower relative thresholds are used for the other languages, reflecting the more limited prevalence of these languages: the thresholds are made relative to the amount of data per language and are comparable to the English and Spanish threshold. For English and Spanish, the national varieties align across both datasets; thus, the experiments for these two languages are paired and we also consider similarity of models across registers. But for the other languages aligning the national varieties in this way removes too many from consideration; thus, there is no cross-domain evaluation for Arabic, French, German, Portuguese, or Russian.

The inventory of national varieties in Table 2 is entirely data-driven and does not depend on distinctions like dialects vs. varieties, inner-circle vs. outer-circle, or native vs. non-native. Instead, the selection is empirical: any area with a large amount of observed English usage is assumed to represent a national variety of English. Since the regions here are based on national boundaries, we call these national varieties. We could just as easily call them national dialects or regional varieties. The global distribution of national varieties for each language is shown in Figure 2.

The datasets are formed into training, testing, and development sets as follows: First, 2k samples are used for development purposes regardless of the amount of data from a given variety. Depending on the size of each variety, at least 12k training and 2.5k testing samples are available. Because some varieties are represented by much larger corpora (i.e., Tweets from American English), a maximum of 25k training samples and 5k testing samples are allowed per variety per register. These datasets contain significantly more observations than have been used in previous work \citep{d18b}.

For each language, we compare six sets of features: First, syntactic representations using CxG-1 and CxG-2; Second, indirect syntactic representations using function words\footnote{For replicability, these are taken from \url{https://github.com/stopwords-iso}}; Third, unigrams and bigrams and trigrams of lexical items. Lexical unigrams represent mostly non-syntactic information while increasing the size of $n$ begins to indirectly include information about transitions. The n-grams are representing using a hashing vectorizer with 30k dimensions (thus, these representations have no syntactic features present). This avoids biasing the selection of specific n-grams (i.e., with content more associated with dominant inner-circle varieties). But this also means that the lexical features themselves cannot be inspected.

\section{Results}

This section reports the results of dialectometry experiments across seven languages. First, in Section 3.1 we look at overall predictive accuracy using the F-Measure metric across feature sets and languages. The purpose of this analysis is to contextualize and then explore the interpretation of classification-based dialectometry. Second, in Section 3.2 we examine the robustness of models across registers (using the web corpus and the Twitter corpus) and across rounds of feature pruning. The purpose of this analysis is to understand how meaningful these models are in the presence of possible confounds such as a reliance on a small number of highly predictive variants. These first two sections are important for validating a classification-based approach to syntactic variation. Third, in Section 3.3 we analyze predictive accuracy and prediction errors across languages and develop representations of regional syntactic similarity. The purpose of this analysis is to use dialect classification to understand global syntactic variation in the aggregate. Fourth, in Section 3.4 we examine measures of the uniqueness of different regional varieties and in Section 3.5 we apply these models to evaluate empirical correlates for notions like \textit{inner-circle} and \textit{outer-circle}. These last two sections are important for understanding what dialect classification can tell us about global, whole-grammar syntactic variation once the approach itself has been validated.

\subsection{Features, Model Size, and Predictive Accuracy}

The overall prediction accuracy across languages is shown in Table 3 (with the web corpus above and the Twitter corpus below). On the left-hand part of the table, the syntactic features are grouped: function words and the two CxG feature sets. On the right-hand part, the lexical features are grouped: lexical unigrams, bigrams, and trigrams. For reference, the number of regions for each variety is shown in the final column. 

A classification-based approach has the goal of distinguishing between national varieties. We would expect, then, that the task of distinguishing between a small number of varieties is easier than distinguishing between a larger number of varieties. For example, there are only two varieties of German and Portuguese in the Twitter corpus. For Portuguese, all feature sets have F1s of 1.00 or 0.99; in other words, this is an easy task and there are many ways of doing it. This is also an indication that these varieties of Portuguese (here, from Brazil, BR, and from Portugal, PT) are quite distinct across all feature sets. On the other hand, even though German also has a small number of national varieties (here, from Germany, DE, and from Austria, AT), there is a wide variation in prediction accuracy, with function words (F1 = 0.83) and CxG-1 (F1 = 0.90) having markedly lower performance than other feature sets. The point is that model performance depends on both the number of national varieties included in the model (showing the importance of taking an empirical approach to the selection of varieties) as well as on the degree of difference between the varieties themselves. Portuguese as used in Brazil and Portugal is significantly more distinct than German as used in Germany and Austria. Digging deeper, however, we also notice that function words as features are more uneven across languages than other feature sets. For example, Arabic on Twitter has eight national varieties and function words achieve an F1 of 0.80; but for Russian on Twitter, with only three varieties, function words achieve a lower F1 of 0.73. This is an indication that, as indirect proxies for syntactic structure, the usefulness of function words for this task varies widely by language (at least, given the inventory of function words used here).

\begin{table}[b]
\begin{center}
	\def\arraystretch{1.5}
	\begin{tabular}{|l|ccc|ccc|c|}
		\hline
\textbf{CC} & \textbf{Function} & \textbf{CxG-1 ~ } & \textbf{CxG-2 ~ } & \textbf{Unigram} & \textbf{Bigram} & \textbf{Trigram} & \textbf{N. Regions}\\
\hline
Arabic & 0.88 & 0.90 & 1.00 & 1.00 & 1.00 & 0.96 & 4 \\
English & 0.65 & 0.80 & 0.96 & 1.00 & 0.98 & 0.87 & 14 \\
French & 0.61 & 0.78 & 0.96 & 1.00 & 0.98 & 0.90 & 13 \\
German & 0.84 & 0.89 & 0.96 & 1.00 & 0.98 & 0.86 & 8 \\
Portuguese & 0.89 & 0.98 & 0.99 & 1.00 & 1.00 & 0.97 & 4 \\
Russian & 0.41 & 0.79 & 0.95 & 1.00 & 0.95 & 0.80 & 19 \\
Spanish & 0.52 & 0.78 & 0.95 & 1.00 & 0.99 & 0.91 & 17 \\
		\hline
\textbf{TW} & \textbf{Function} & \textbf{CxG-1 ~ } & \textbf{CxG-2 ~ } & \textbf{Unigram} & \textbf{Bigram} & \textbf{Trigram} & \textbf{N. Regions}\\
\hline
Arabic & 0.80 & 0.88 & 0.98 & 1.00 & 1.00 & 0.94 & 8 \\
English & 0.55 & 0.76 & 0.92 & 1.00 & 0.97 & 0.82 & 14 \\
French & 0.88 & 0.98 & 0.98 & 1.00 & 1.00 & 0.99 & 4 \\
German & 0.83 & 0.90 & 0.95 & 1.00 & 0.99 & 0.95 & 2 \\
Portuguese & 1.00 & 0.99 & 1.00 & 1.00 & 1.00 & 0.99 & 2 \\
Russian & 0.73 & 0.83 & 0.93 & 1.00 & 0.94 & 0.87 & 3 \\
Spanish & 0.51 & 0.82 & 0.94 & 1.00 & 0.99 & 0.92 & 17 \\
		\hline
	\end{tabular}
	\caption{F1 of Classification of Regional Varieties by Language and Feature Type \\ (Web Corpus Above and Twitter Corpus Below)}
	\label{tab:1}
	\end{center}
\end{table}

Regardless of the number of national varieties per language, lexical unigrams perform the best (F1 = 1.00). In other words, it is not difficult to disinguish between samples from New Zealand and Australia when given access to lexical items (\textit{Christchurch} vs. \textit{Brisbane}). While we know that syntactic models are capturing linguistic variation, however, the success of lexical models, as argued elsewhere \citep{Dunn2019a}, is partly a result of place-names, place-specific content, and place-specific entities. In other words, geo-referenced texts capture the human geography of particular places and this human geography information takes the form of specific lexical items. Previous work has focused on capturing precisely this type of content \citep{Wing2014,hsf15,Adams2015,lmz17,Adams2018}. The problem is that, without organizing the frequency of such lexical features according to concept \citep{Zenner2012}, these models may not represent linguistic variation.

This is a simplification, of course, but the underlying point is that it is difficult to distinguish linguistic lexical variation from human geography-based and topical lexical variation without relying on the idea of conceptual alternations. For example, we know that as \textit{n} increases n-grams represent increasing structural information (i.e., transitions between lexical items instead of lexical items in isolation). Here we see that, by the time \textit{n} is raised to three, the predictive accuracy of CxG-2 always surpasses the predictive accuracy of trigrams (with the single exception of French on Twitter). The difference between CxG-2 and bigrams is much smaller than the distance between the various syntactic features. This is evidence that the advantage of unigrams over CxG-2 reflects the advantage of human geography content (i.e., lexical items in isolation) over linguistic variation (i.e., transitions between lexical items). In short, while some of the lexical variation is linguistic (\textit{soda} vs. \textit{pop}), a good deal of it is also based on human geography (\textit{Chicago} vs. \textit{Singapore}). The advantage of syntactic models in this context is that such non-linguistic variations do not introduce confounds: we know that these models represent regional varieties of each language.

Models on the web corpus (above) have higher predictive accuracy than models on the Twitter corpus (below). This is true except in cases, such as Portuguese, where there is a wide difference in the number of national varieties represented (for Portuguese, two vs. four). For reasons of data availability, only English and Spanish have strictly aligned varieties; in both of these languages, the syntactic features perform better on the web corpus than the Twitter corpus, although the gap is wider for English than for Spanish. This raises a question that is addressed in the next section: are models of syntactic variation consistent across these registers? In other words, do the web-based and Twitter-based models make the same types of errors?

The web corpus also provides more varieties per language (with Arabic as the sole exception, which is better represented on Twitter). In many cases this difference is significant: there are 19 varieties of Russian on the web, but only three on Twitter. In this case, there are competing Russian-language social media platforms (i.e., \url{www.vk.com}) that are not included in this study. In other words, outside of English and Spanish, which are aligned across datasets, the Twitter data is less comprehensive.

\begin{figure}
\begin{center}
    \textbf{Web}
	\includegraphics[scale=1.0]{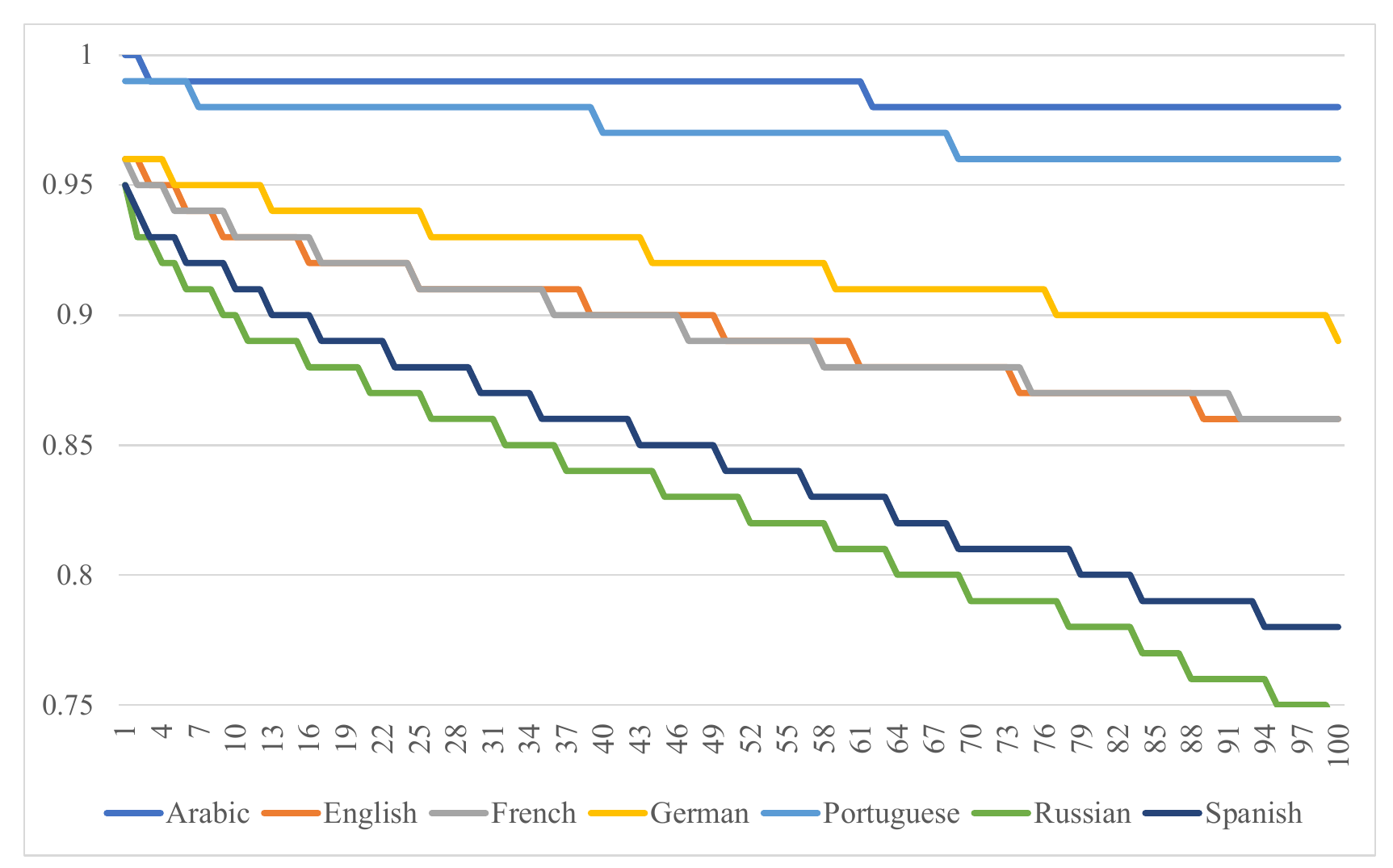}
	\textbf{Twitter}
	\includegraphics[scale=1.0]{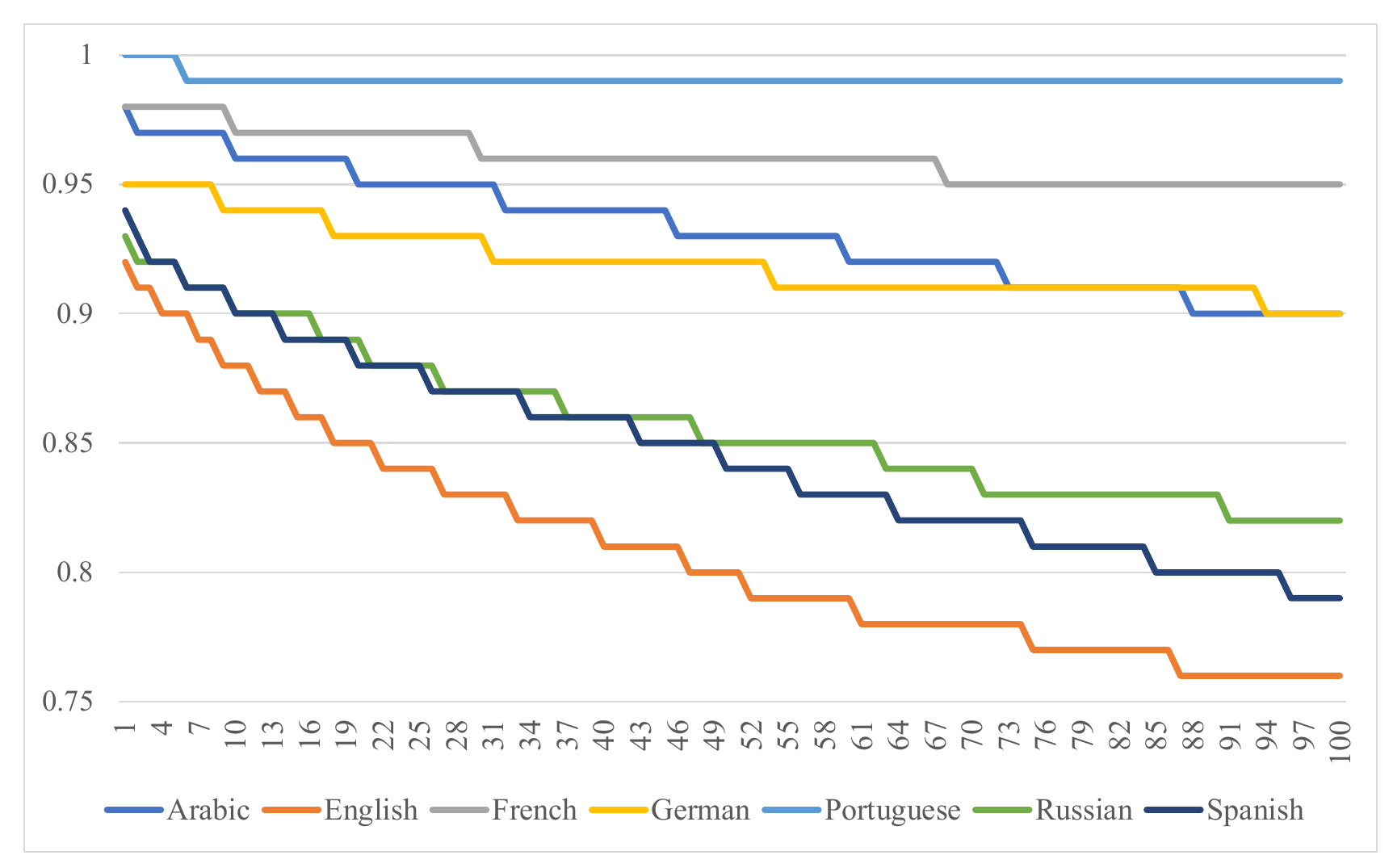}
	\caption{Model Robustness by Language Using Unmasking for 100 Iterations with CxG-2 Features \\
	(Web Models Above and Twitter Models Below)}
	\end{center}
\end{figure}

What does the F-Measure tell us about models of syntactic variation? First, the measure is a combination of precision and recall that reflects the predictive accuracy while taking potentially imbalanced classes into account: how many held-out samples can be correctly assigned to their actual region-of-origin? On the one hand, this is a more rigorous evaluation than simply finding a significant difference in a syntactic feature across varieties within a single-fold experimental design: not only is there a difference in the usage of a specific feature, but we can use the features in the aggregate to characterize the difference between national varieties. On the other hand, it is possible that a classifier is over-fitting the training data so that the final model inflates the difference between varieties. For example, let's assume that there is a construction that is used somewhat frequently in Pakistan English but is never used in other varieties. In this case, the classifier could achieve a very high prediction accuracy while only a single construction is actually in variation. Before we interpret these models further, the next section evaluates whether this sort of confound is taking place.

\subsection{Model Robustness Across Features and Registers}

If a classification model depends on a small number of highly predictive features, thus creating a confound for dialectometry, the predictive accuracy of that model will fall abruptly as such features are removed \citep{ksb07}. Within authorship verification, \textit{unmasking} is used to evaluate the robustness of a text classifier: First, a linear classifier is used to separate documents; here, a Linear SVM is used to classify national varieties of a language. Second, for each round of classification, the features that are most predictive are removed: here, the highest positive and negative features for each national variety are pruned from the model. Third, the classifier is retrained without these features and the change in predictive accuracy is measured: here, unmasking is run for 100 iterations using the CxG-2 grammar as features, as shown in Figure 3 (with the web-based model above and the Twitter-based model below). For example, this removes 28 constructions from the model of English each iteration (two for each national dialect), for a total of approximately 2,800 features removed. The figures show the F-Measure for each iteration. On the left-hand side, this represents the performance of the models with all features are present; on the right-hand side, this represents the performance of the models after many features have been removed. This provides a measure of the degree to which these models are subject to a few highly predictive features.

First, we notice that models with a higher starting predictive accuracy (e.g., Arabic and Portuguese in the web-based model and Portuguese and French in the Twitter-based model) tend to maintain their accuracy across the experiment. Even after 100 rounds of pruning, Arabic and Portuguese (CC) remain above 0.95 with CxG-2 features.\footnote{Here and below we focus on CxG-2 as the highest performing syntactic model.} Similarly, French and Portuguese remain above 0.95 after 100 rounds of pruning (TW). This indicates that a high performing dialect classification model is based on a broad and distributed set of features. But this is not always the case: for example, Arabic (TW) starts out with the same performance as French but over the course of the experiment declines to a performance that is 10\% lower than French. This is an indication that this Twitter-based model of Arabic is less robust than its counter-part model of French (although keep in mind that the French model has only 4 varieties and the Arabic model has 8).

Second, although Spanish and Russian have a starting accuracy that is comparable to other languages, with F1s of 0.95 for both languages on the web corpus, their accuracy falls much more quickly. Spanish and Russian decrease by around 20\% by the end of the experiment while English and French decrease by only 10\% in total. On the Twitter corpus, Spanish and Russian again pattern together, this time with a 15\% reduction. But here the English model has a somewhat steeper decline. In most cases, however, the starting accuracy of a model is related to its rate of decline: more accurate models are also more robust to feature pruning. The purpose of this evaluation is to show that a classification approach to dialectometry is not subject to the confound of a small number of highly predictive features.

\begin{figure*}
\begin{center}
    \textbf{English}
	\includegraphics[scale=0.97]{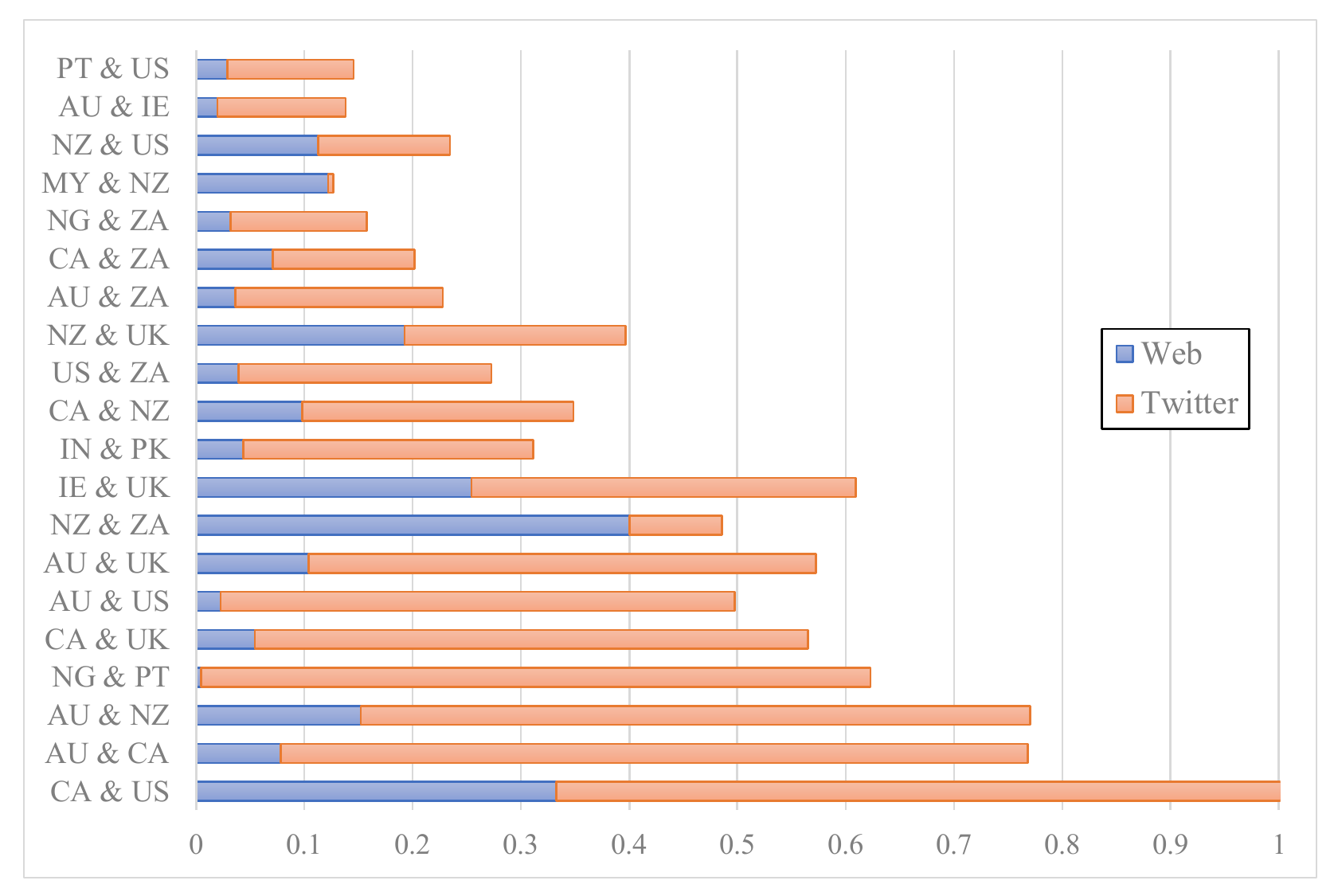}
	\textbf{Spanish}
	\includegraphics[scale=0.97]{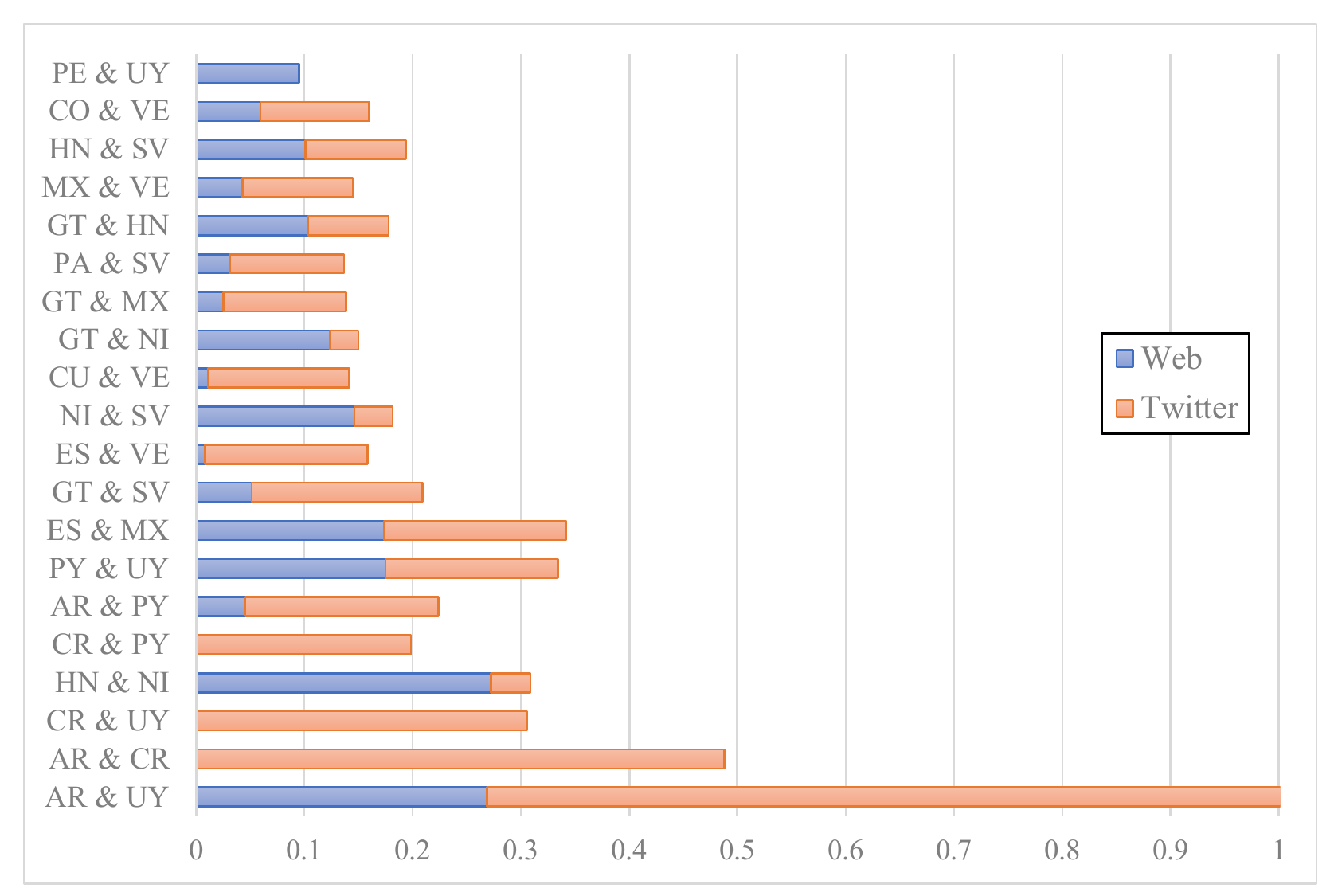}
	\caption{Classification Errors by Percent of Dataset for Web and Twitter Corpora using CxG-2 Features \\ (English Errors Above and Spanish Errors Below)}
	\end{center}
\end{figure*}

The next question is about the similarity of national varieties as represented in the web corpus vs. the Twitter corpus. Is there a consistent representation of variation or are the models ultimately register-specific? For this analysis we focus on English and Spanish as the two languages that are aligned by national varieties across both datasets. We focus on an analysis of errors: First, two national varieties that are more often confused by the classifier are more similar according to the model. Thus, we represent the similarity of regions using the total of all errors between two varieties. For example, if UK English is predicted to be New Zealand English 50 times and New Zealand English is predicted to be UK English 25 times, there are 75 total errors between these varieties. More errors reflects more similar varieties.\footnote{Country abbreviations are given in Appendix A.}

The question is whether the web corpus and Twitter both provide the same patterns of similarity. Figure 4 shows the relative errors between varieties for both datasets (with English above and Spanish below): the web (blue) occupies the left-hand side of each bar and Twitter (red) occupies the right-hand side. If both colors are the same size, we see the same proportion of errors for a given pair across both datasets. This figure also shows the most similar varieties, with the varieties having the highest total errors occupying the bottom of each. For example, the most similar varieties of English on Twitter are American (US) and Canadian English (CA). The most similar varieties on the web corpus, however, are New Zealand (NZ) and South African English (ZA).\footnote{The ISO country codes are used in all figures and tables; these are shown by common name in the first Appendix.} The Pearson correlation between errors, paired across datasets by varieties, is highly significant for English at 0.494 (note that this involves the number of errors but does not require that the errors themselves match up across registers). At the same time, there remain meaningful differences between the datasets. For example, Nigeria (NG) and Portugal (PT) have many errors in the Twitter model but very few in the web model. On the other hand, New Zealand (NZ) and South Africa (ZA) have a large number of errors in the web model but few in the Twitter model. This is an indication that the models are somewhat different across registers.

The errors for Spanish, in the bottom portion of Figure 4, also are significantly correlated across registers, although the Pearson correlation is somewhat lower (0.384). For example, both corpora have significant errors between Argentina (AR) and Uruguay (UY), although Twitter has a much higher error rate. But errors between Costa Rica (CR) and Uruguay (UY) and between Argentina (AR) and Costa Rica (CR) are only found on Twitter. Errors between Honduras (HN) and Nicaragua (NI), on the other hand, are only found in the web model. The point is that the two registers are associated in their error rates for both English and Spanish (the only languages with regional varieties aligned across both datasets).

The high accuracy of these models could suggest that the models are over-fitting the test set, even with a relatively large number of samples in the test set. Thus, in Table 4, we compare the weighted F1 scores on the test set with a 10-fold cross-validation evaluation that includes the training and testing data together. The table shows the maximum and minimum values across folds. There are only three cases in which the minimum fold F1 is lower than the reported test set metrics: Russian (web data), Arabic (Twitter data), and Portuguese (Twitter data). In each case the difference is small and in each case the average fold F1 is the same as the F1 from the test set alone. This evidence shows that the models are not over-fitting the test set and that this reflects a robust classification accuracy.

\begin{table}
\begin{center}
	\def\arraystretch{1.5}
	\begin{tabular}{|c|ccc|c|ccc|}
		\hline
		~ & ~ & \textbf{CC} & ~ & ~ & ~ & \textbf{TW} & ~ \\
		\hline
		~ & \textbf{Train-Test} & \textbf{CV-Max} & \textbf{CV-Min} & ~ & \textbf{Train-Test} & \textbf{CV-Max} & \textbf{CV-Min} \\
		\hline
		Arabic & 1.00 & 1.00 & 1.00 & Arabic & 0.98 & 0.98 & \textbf{0.97} \\
        English & 0.96 & 0.96 & 0.96 & English & 0.92 & 0.92 & 0.92 \\
        French & 0.96 & 0.96 & 0.96 & French & 0.98 & 0.98 & 0.99 \\
        German & 0.96 & 0.96 & 0.96 & German & 0.95 & 0.96 & 0.95 \\
        Portuguese & 0.99 & 0.99 & 0.99 & Portuguese & 1.00 & 1.00 & \textbf{0.99} \\
        Russian & 0.95 & 0.95 & \textbf{0.94} & Russian & 0.93 & 0.95 & 0.93 \\
        Spanish & 0.95 & 0.95 & 0.95 & Spanish & 0.94 & 0.94 & 0.94 \\
		\hline
	\end{tabular}
	\caption{Comparison of Train/Test and Cross-Validation Results by Weighted F1 for CxG-2}
	\label{tab:1}
	\end{center}
\end{table}

This section has approached two important questions: First, is a classification model dependent on a small number of highly predictive features? Second, does a classification model produce the same type of errors across both web corpora and Twitter corpora? In both cases some languages (like English) are more robust across feature pruning and more stable across registers than others (like Spanish). This is the case even though the F-Measure (reflecting predictive accuracy alone) is similar for both languages: 0.96 vs. 0.95 for the web model and 0.92 vs. 0.94 for the Twitter model. These alternate evaluations, then, are important for revealing further properties of these classification models. The predictive accuracy for both languages is high across both registers and the regional varieties which are confused is significantly correlated across both registers.

\begin{table}
\begin{center}
	\def\arraystretch{1.5}
	\begin{tabular}{|l|ccc|c|ccc|}
		\hline
~ & \textbf{Prec (CC)} & \textbf{Recall (CC)} & \textbf{F1 (CC)} & ~ & \textbf{Prec (TW)} & \textbf{Recall (TW)} & \textbf{F1 (TW)} \\
\hline
AU & 0.97 & 0.96 & 0.97 & AU & 0.82 & 0.83 & 0.83 \\
CA & 0.94 & 0.94 & 0.94 & CA & 0.84 & 0.79 & 0.81 \\
IE & 0.97 & 0.97 & 0.97 & IE & 0.95 & 0.95 & 0.95 \\
NZ & 0.91 & 0.92 & 0.91 & NZ & 0.92 & 0.90 & 0.91 \\
UK & 0.95 & 0.95 & 0.95 & UK & 0.87 & 0.90 & 0.89 \\
US & 0.93 & 0.95 & 0.94 & US & 0.85 & 0.89 & 0.87 \\
ZA & 0.94 & 0.96 & 0.95 & ZA & 0.92 & 0.94 & 0.93 \\
\hline
IN & 0.97 & 0.98 & 0.97 & IN & 0.97 & 0.97 & 0.97 \\
MY & 0.96 & 0.96 & 0.96 & MY & 0.99 & 0.99 & 0.99 \\
NG & 0.98 & 0.98 & 0.98 & NG & 0.94 & 0.95 & 0.94 \\
PH & 0.98 & 0.97 & 0.98 & PH & 0.98 & 0.98 & 0.98 \\
PK & 1.00 & 0.99 & 0.99 & PK & 0.98 & 0.98 & 0.98 \\
\hline
CH & 0.97 & 0.94 & 0.96 & CH & 0.98 & 0.97 & 0.97 \\
PT & 0.99 & 0.98 & 0.98 & PT & 0.93 & 0.90 & 0.92 \\
\hline
\textbf{AVG} & \textbf{0.96} & \textbf{0.96} & \textbf{0.96} & \textbf{AVG} & \textbf{0.92} & \textbf{0.92} & \textbf{0.92} \\
		\hline
	\end{tabular}
	\caption{Classification Performance for English Regions, Web and Twitter Corpora, CxG-2 Features}
	\label{tab:1}
	\end{center}
\end{table}

\subsection{Regional Accuracy and Similarity}

While the previous sections have evaluated classification-based models externally (prediction accuracy by feature type, robustness across feature pruning, error similarity across registers), this section and the next focus on internal properties of the models: what are the relationships between national varieties for each language? Which regions perform best within a model? In this section we examine the F-Measure of individual national varieties and the similarity between varieties using cosine similarity between feature weights. Because the Twitter dataset has fewer varieties for most languages, we focus on similarity within the web models alone and only for languages with a large inventory of varieties (i.e., only for English, French, and Spanish).

\begin{figure*}
\begin{center}
	\includegraphics[scale=0.99]{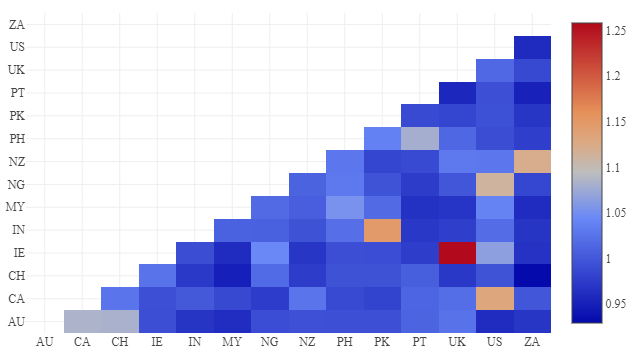}
	\caption{Region Similarity by Cosine Between Feature Weights, English CxG-2}
	\end{center}
\end{figure*}

\begin{table}
\begin{center}
	\def\arraystretch{1.5}
	\begin{tabular}{|l|ccc|c|ccc|}
		\hline
~ & \textbf{Prec (CC)} & \textbf{Recall (CC)} & \textbf{F1 (CC)} & ~ & \textbf{Prec (TW)} & \textbf{Recall (TW)} & \textbf{F1 (TW)} \\
\hline
BE & 0.94 & 0.86 & 0.90 & BE & 0.97 & 0.94 & 0.96 \\
BF & 0.98 & 0.98 & 0.98 & BF & -- & -- & -- \\
CH & 0.92 & 0.93 & 0.93 & CH & -- & -- & -- \\
CM & 1.00 & 1.00 & 1.00 & CM & -- & -- & -- \\
DZ &  0.99 & 0.99 & 0.99 & DZ & -- & -- & -- \\
FR & 0.92 & 0.95 & 0.93 & FR & 0.97 & 0.98 & 0.98 \\
GD & 0.94 & 0.92 & 0.93 & GD & -- & -- & -- \\
HT & -- & -- & -- & HT & 1.00 & 1.00 & 1.00 \\
LU & 0.97 & 0.96 & 0.96 & LU & 1.00 & 1.00 & 1.00 \\
NC & 0.96 & 0.95 & 0.95 & NC & -- & -- & -- \\
PF & 0.97 & 0.97 & 0.97 & PF & -- & -- & -- \\
RE & 0.94 & 0.95 & 0.95 & RE & -- & -- & -- \\
SN & 0.98 & 0.98 & 0.98 & SN & -- & -- & -- \\
TN & 0.98 & 0.97 & 0.98 & TN & -- & -- & -- \\
\hline
\textbf{AVG} & \textbf{0.96} & \textbf{0.96} & \textbf{0.96} & \textbf{AVG} & \textbf{0.98} & \textbf{0.98} & \textbf{0.98} \\
		\hline
	\end{tabular}
	\caption{Classification Performance for French Regions, Web and Twitter Corpora, CxG-2 Features}
	\label{tab:1}
	\end{center}
\end{table}

\begin{figure*}
\begin{center}
    \textbf{French}
	\includegraphics[scale=0.99]{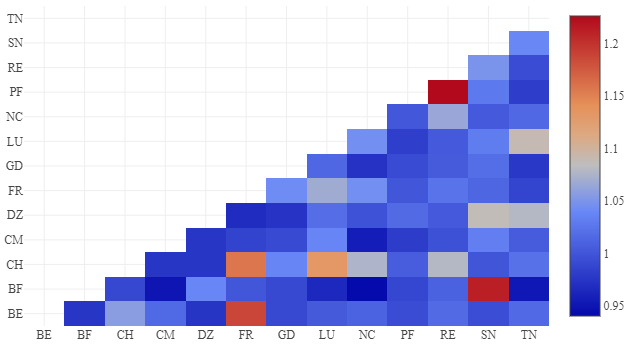}
	\textbf{Spanish}
	\includegraphics[scale=0.99]{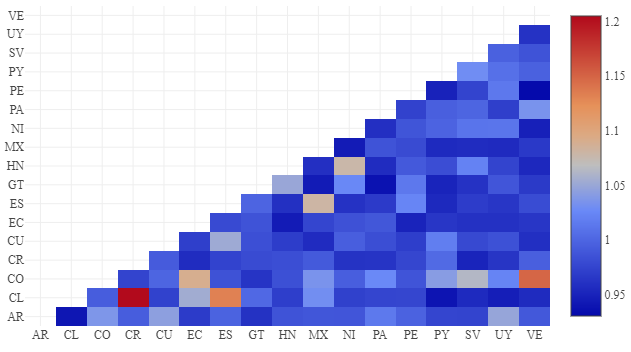}
	\caption{Region Similarity by Cosine Between Feature Weights, French (Above) and Spanish (Below) CxG-2}
	\end{center}
\end{figure*}

We start with English in Table 5. The left-hand side shows Precision, Recall, and F-Measure scores for the web corpus and the right-hand side for the Twitter corpus, both using the CxG-2 feature set. The higher the scores for each national dialect, the more distinct that variety is from the others in syntactic terms. New Zealand English (NZ) has the lowest F1 (0.91) for the web corpus. While the score of NZ English is the same for the Twitter model (0.91), it is no longer the lowest scoring variety: this is now Canadian English (CA) at 0.81. In fact, the lowest performing varieties for the Twitter model are all \textit{inner-circle} varieties: Australia (AU), Canada (CA), United Kingdom (UK), and the United States (US). This phenomenon is explored further in the next section: why are more dominant varieties more difficult to model? Is this consistent across languages? For now we note only that all of the countries included in the model are expected, with perhaps the exception of Portugal (PT) and Switzerland (CH). While previous work made an explicit distinction between inner-circle and outer-circle varieties \citep{d18b}, here we leave this type of categorization as an empirical question.

We can compare national varieties within a model by comparing their respective feature weights: which regions have the most similar syntactic profiles? We use cosine distance to measure the similarity between feature weights and then use a heat map, as in Figure 5, to visualize the similarities. Cells with a higher value (more red) indicate a pair of varieties which the model is trying hard to separate (thus, a more similar pair). For example, the most similar pair is UK English (UK) and Irish English (IE); this is expected given that Northern Ireland is part of the UK. The next four pairs also are expected: Indian (IN) and Pakistan English (PK), American (US) and Canadian English (CA), New Zealand (NZ) and South African English (ZA), American (US) and Nigerian English (NG). While the final pair is less transparent, it is important that the model picks out these pairs of related varieties without any pre-knowledge. On the other hand, dark blue values indicate that the model is not concerned with separating the pair (because they are not very similar): for example, South African English (ZA) and Swiss English (CH).

French varieties are shown in Table 6, with again a much larger inventory for the web model than for the Twitter model. As with English, the lowest performing varieties in terms of prediction accuracy are the most dominant inner-circle varieties: France (FR), Belgium (BE), and Switzerland (CH). One possible reason is that there is more internal variation in France than in, for example, Cameroon (CM). Another possible reason is that these inner-circle varieties have influenced the outer-circle varieties, so that they are harder to distinguish from the colonial varieties. The regions in the web model are expected given French colonial history: European varieties (France, Switzerland, Belgium, Luxembourg), African varieties (Burkina Faso, Cameroon, Senegal), North African varieties (Grenada, Algeria, Tunisia), Pacific varieties (New Caledonian, French Polynesia), and unconnected island varieties with current or past French governance (R\'{e}union, Grenada). All have a history of French usage.

\begin{table}
\begin{center}
	\def\arraystretch{1.5}
	\begin{tabular}{|l|ccc|c|ccc|}
		\hline
~ & \textbf{Prec (CC)} & \textbf{Recall (CC)} & \textbf{F1 (CC)} & ~ & \textbf{Prec (TW)} & \textbf{Recall (TW)} & \textbf{F1 (TW)} \\
\hline
AR & 0.94 & 0.94 & 0.94 & AR & 0.85 & 0.90 & 0.87 \\
CL & 0.99 & 0.98 & 0.98 & CL & 0.97 & 0.98 & 0.97 \\
CO & 0.95 & 0.94 & 0.95 & CO & 0.95 & 0.93 & 0.94 \\
CR & 1.00 & 1.00 & 1.00 & CR & 0.91 & 0.87 & 0.89 \\
CU & 0.96 & 0.97 & 0.97 & CU & 0.98 & 0.97 & 0.98 \\
EC & 0.96 & 0.96 & 0.96 & EC & 0.98 & 0.98 & 0.98 \\
ES & 0.94 & 0.95 & 0.94 & ES & 0.94 & 0.96 & 0.95 \\
GT & 0.96 & 0.96 & 0.96 & GT & 0.94 & 0.95 & 0.95 \\
HN & 0.93 & 0.94 & 0.94 & HN & 0.94 & 0.92 & 0.93 \\
MX & 0.94 & 0.93 & 0.93 & MX & 0.92 & 0.93 & 0.93 \\
NI & 0.92 & 0.86 & 0.89 & NI & 0.98 & 0.98 & 0.98 \\
PA & 0.98 & 0.98 & 0.98 & PA & 0.95 & 0.95 & 0.95 \\
PE & 0.94 & 0.92 & 0.93 & PE & -- & -- & -- \\
PY & 0.94 & 0.96 & 0.95 & PY & 0.93 & 0.94 & 0.93 \\
SV & 0.95 & 0.94 & 0.95 & SV & 0.93 & 0.94 & 0.93 \\
UY & 0.91 & 0.93 & 0.92 & UY & 0.88 & 0.85 & 0.86 \\
VE & 0.97 & 0.98 & 0.98 & VE & 0.94 & 0.93 & 0.93 \\
\hline
\textbf{AVG} & \textbf{0.95} & \textbf{0.95} & \textbf{0.95} & \textbf{AVG} & \textbf{0.94} & \textbf{0.94} & \textbf{0.94} \\
		\hline
	\end{tabular}
	\caption{Classification Performance for Spanish Regions, Web and Twitter Corpora, CxG-2 Features}
	\label{tab:1}
	\end{center}
\end{table}

Following the same methodology for English, region similarity is shown in Figure 6. The closest varieties are from R\'{e}union and French Polynesia, from Senegal and Burkina Faso, and from France and Belgium. This again shows that the model not only distinguishes between varieties but can also situate the varieties in relationship to one another.

Next, regional accuracies for Spanish are shown in Table 7; these are aligned by country with the exception of Peru (PE) which is missing from the Twitter dataset. There is a single European variety (Spain), South American varieties (Argentina, Chile, Colombia, Ecuador, Peru, Paraguay, Uruguay, Venezuela), Central American varieties (Costa Rica, Guatemala, Honduras, Nicaragua, Panama, El Salvador), as well as Cuban and Mexican varieties. The alignment across datasets helps to ensure that only expected varieties occur; as discussed above, there is in fact a significant correlation between the errors produced on the two datasets.

The similarity between Spanish regions is shown in Figure 6 (below French). The most similar varieties are from Costa Rica and Chile, from Spain and Chile, and from Venezuela and Colombia. The least similar are from Argentina and Chile and from Peru and Venezuala.

\begin{table}[b]
\begin{center}
	\def\arraystretch{1.5}
	\begin{tabular}{|l|ccc|c|ccc|}
		\hline
~ & \textbf{Prec (CC)} & \textbf{Recall (CC)} & \textbf{F1 (CC)} & ~ & \textbf{Prec (TW)} & \textbf{Recall (TW)} & \textbf{F1 (TW)} \\
\hline
AZ & 0.94 & 0.94 & 0.94 & AZ & -- & -- & -- \\
BG & 1.00 & 1.00 & 1.00 &  BG & -- & -- & -- \\
BY & 0.98 & 0.95 & 0.97 & BY & 0.91 & 0.85 & 0.88 \\
EC & 0.96 & 0.98 & 0.97 &  EC & -- & -- & -- \\
EE & 0.86 & 0.89 & 0.87 &  EE & -- & -- & -- \\
GE & 0.95 & 0.95 & 0.95 &  GE & -- & -- & -- \\
HT & 0.99 & 0.99 & 0.99 &  HT & -- & -- & -- \\
KG & 0.99 & 0.99 & 0.99 &  KG & -- & -- & -- \\
KZ & 0.96 & 0.93 & 0.94 &  KZ & -- & -- & -- \\
LT & 0.94 & 0.93 & 0.94 &  LT & -- & -- & -- \\
LV & 0.92 & 0.91 & 0.91 &  LV & -- & -- & -- \\
MD & 0.98 & 0.97 & 0.97 &  MD & -- & -- & -- \\
RU & 0.90 & 0.90 & 0.90 & RU & 0.93 & 0.96 & 0.94 \\
SI & 1.00 & 1.00 & 1.00 &  SI & -- & -- & -- \\
TJ & 0.95 & 0.97 & 0.96 &  TJ & -- & -- & -- \\
UA & 0.93 & 0.94 & 0.94 & UA & 0.98 & 0.96 & 0.97 \\
UZ & 0.92 & 0.92 & 0.92 &  UZ & -- & -- & -- \\
\hline
\textbf{AVG} & \textbf{0.95} & \textbf{0.95} & \textbf{0.95} & \textbf{AVG} & \textbf{0.94} & \textbf{0.94} & \textbf{0.93} \\
		\hline
	\end{tabular}
	\caption{Classification Performance for Russian Regions, Web and Twitter Corpora, CxG-2 Features}
	\label{tab:1}
	\end{center}
\end{table}

Russian varieties are shown in Table 8, encompassing much of Eastern Europe and Central Asia. As mentioned before, the Twitter dataset is missing a number of important varieties, most likely because of the influence of other social media platforms. There are two noisy regions, SO and PW, present in the web corpus.\footnote{One approach that could remove the few noisy regions that show up in Russian and, later, in German is to use population-based sampling to reduce the amount of data per country before selecting regional varieties.} Beyond this, the countries represented are all expected: in addition to Russia (RU), there are varieties from Central Asia (Azerbaijan, Georgia, Kyrgyzstan, Tajikistan, Uzbekistan), Southeast Europe (Bulgaria, Moldova), and Eastern Europe (Belarus, Lithuania, Slovenia, Ukraine). There are also varieties that reflect expanding-circle varieties of Russian (Ecuador, Haiti). Given the lack of alignment between the datasets, it is difficult to evaluate whether or not these expanding-circle varieties are robust. This reflects another limitation of an entirely data-driven approach: when is the use of Russian in a country a stable dialect and when is it a non-native variety that reflects short-term military or economic connections? The capacity of this syntactic model to predict both suggests that, in empirical terms, the distinction is not important. It could be the case, however, that some varieties are more robust than others to feature pruning. For reasons of space, similarities between Russian varieties are not shown.

Because they have fewer national varieties each, we end with Arabic, German, and Portuguese together (this table is shown in Appendix 2). Starting with Arabic, the regional comparison is made difficult by the little overlap between the two datasets: only data from Syria is consistent across registers. Focusing on the Twitter model, then, we note that it does contain examples of several traditional dialect groups: Algerian (DZ) represents the Maghrebi group, Egypt (EG) represents the Egyptian group, Iraq (IQ) and Syria (SY) represent the Mesopotamian group, Jordan (JO) and Palestine (PS) represent the Levantine group, and Kuwait (KW) represents the Arabian group. In addition, there is a Russian (RU) dialect of Arabic, reflecting an emerging outer-circle variety. Given the sparsity of regions shared across the two datasets, we do not explore further the relationships between varieties. The point here is to observe that the models on both datasets maintain a high accuracy across regions and that the available countries do represent many traditional dialect groups.

For German, Twitter provides only a few inner-circle varieties. Here we see, again, that the most central or proto-typical dialect (Germany, DE) has the lowest overall performance while the highest performance is found in less-central varieties. While other languages have national varieties representing countries that we expect to see, the German web corpus contains three regions that are almost certainly noise: the PW (Palau), SO (Somalia), and TL (East Timor) domains are most likely not used for regional web pages but rather for other purposes. No other language has this sort of interference by non-geographic uses of domain names (except that Russian also picks up data from \textit{.so} and \textit{.pw}). Most likely this results from having a frequency threshold that is too low. Because a classifier attempts to distinguish between all classes, the inclusion of noisy classes like this may reduce performance but will never improve performance. Thus, we leave this model as-is in order to exemplify the sorts of problems that an entirely data-driven methodology can create. Ignoring these varieties, however, the web-based model does provide a well-performing model of Austria (AU), Switzerland (CH), Germany (DE), Luxembourg (LU), and Poland (PL).

For Portuguese, again the Twitter model only covers major varieties: Brazil and Portugal. The web corpus, unlike German, does not show any noisy regions but it does include two expected African varieties: Angola (AO) and Cabo Verde (CV). While the model performs well, we will not delve more deeply into the region-specific results.

The purpose of this section has been to examine the prediction accuracies across national varieties alongside the similarity between varieties. With the exception of some noisy regions for German and Russian, these results show that the model both is able to make accurate predictions about syntactic variation as well as to make reasonable representations of the aggregate similarity between national varieties.

\subsection{Empirical Measures of Region Uniqueness}

We have seen in the sections above that outer-circle or expanding-circle varieties often have higher predictive accuracies even though they are less proto-typical and less dominant. For example, these sorts of varieties have been shown to have lower feature densities for these CxG grammars \citep{Dunn2019a}, which indicates that the grammars are missing certain unique constructions. Regardless, these varieties remain unique in that they are easier to distinguish from more central varieties.

For example, the English Twitter models show the main inner-circle varieties as having the lowest F1 scores: Australia (0.83), Canada (0.81), United States (0.87), and the United Kingdom (0.89). This phenomenon is not limited to English, however. In the French web model, again the inner-circle (i.e., European) varieties have the lowest F1 scores: Belgium (0.90), Switzerland (0.93), and France (0.93). The other languages do not present examples as clear as this; for example, Arabic and German and Portuguese do not contain enough varieties to make such a comparison meaningful. Russian and Spanish are characterized by a large number of varieties that are contiguous in relatively dense regions, thus showing a less striking colonial pattern. Why is it that, in cases of non-contiguous dialect areas, the inner-circle varieties have the lowest prediction accuracy?

In qualitative terms, there are several possible explanations. First, it could be the case that these inner-circle varieties have strongly influenced the other varieties so that parts of their syntactic profiles are replicated within the other varieties. Second, it could be that there is an immigration pipeline from outer-circle to inner-circle countries, so that the samples of UK English, for example, also contain speakers of Nigerian English. Third, it could be the case that media and communications are centered around inner-circle markets so that outer-circle varieties are influenced by one or another center of power. Additional factors could include the strength of standardization across languages, the number of L1 vs L2 speakers that are represented for each language, and the average level of education for each country. None of these possibilities can be distinguished in empirical terms within the current study.

\begin{table}
\begin{center}
	\def\arraystretch{1.5}
	\begin{tabular}{|cccc|}
\hline
\textbf{ ~ ~ ~ English ~ ~ } & \textbf{ ~ ~ ~ French ~ ~ ~ ~ } & \textbf{ ~ ~ ~ Russian ~ ~ ~ } & \textbf{ ~ ~ ~ Spanish ~ ~ } \\
\hline
\end{tabular}
\begin{tabular}{|ccc|ccc|ccc|ccc|}
\hline
1 & US & -0.46 & 1 & FR & -0.49 & 1 & TJ & 0.04 & 1 & ES & -0.24 \\
2 & UK & -0.25 & 2 & RE & -0.36 & 2 & EE & 0.15 & 2 & PY & -0.05 \\
3 & CA & -0.22 & 3 & CH & -0.32 & 3 & SI & 0.17 & 3 & UY & -0.04 \\
4 & NZ & -0.18 & 4 & LU & -0.26 & 4 & LT & 0.23 & 4 & AR & -0.02 \\
5 & AU & -0.16 & 5 & PF & -0.18 & 5 & EC & 0.23 & 5 & CO & 0.00 \\
6 & IE & -0.14 & 6 & SN & -0.15 & 6 & KZ & 0.23 & 6 & CL & 0.03 \\
7 & PH & -0.06 & 7 & BE & -0.10 & 7 & UA & 0.23 & 7 & HN & 0.04 \\
8 & MY & -0.05 & 8 & NC & -0.08 & 8 & LV & 0.26 & 8 & CU & 0.06 \\
9 & IN & -0.01 & 9 & BF & -0.03 & 9 & GE & 0.32 & 9 & MX & 0.10 \\
10 & NG & 0.02 & 10 & GD & -0.02 & 10 & HT & 0.35 & 10 & NI & 0.12 \\
11 & CH & 0.05 & 11 & TN & 0.05 & 11 & KG & 0.35 & 11 & GT & 0.13 \\
12 & ZA & 0.06 & 12 & DZ & 0.07 & 12 & UZ & 0.36 & 12 & SV & 0.15 \\
13 & PT & 0.13 & 13 & CM & 0.25 & 13 & AZ & 0.36 & 13 & CR & 0.18 \\
14 & PK & 0.14 & -- & -- & -- & 14 & BY & 0.47 & 14 & VE & 0.19 \\
-- & -- & -- & -- & -- & -- & 15 & RU & 0.56 & 15 & EC & 0.23 \\
-- & -- & -- & -- & -- & -- & 16 & MD & 0.67 & 16 & PE & 0.25 \\
-- & -- & -- & -- & -- & -- & 17 & BG & 0.84 & 17 & PA & 0.32 \\
\hline
	\end{tabular}
	\caption{Variety Uniqueness by Language using Spearman Correlation, Web CxG-2 Model}
	\label{tab:1}
	\end{center}
\end{table}

We have shown above, however, that this approach to dialectometry can (i) make accurate predictions about variety membership and (ii) can create reasonable representations of aggregate syntactic similarity between regions. In this section we formulate an approach to identifying, in purely synchronic terms, which varieties within a model represent central inner-circle countries that are the sources of influence for other outer-circle countries. The observations about prediction accuracy depend on the evaluation of the model, but we want this measure of uniqueness to depend on the model of variation itself.

The feature weights represent the positive and negative importance of each syntactic feature for each national variety. We used cosine similarities between feature weights above to find the most similar regions. Here we are interested in the overall uniqueness of a particular dialect: which varieties are in general not similar to any other varieties? We calculate this by summing the Spearman correlations between each variety and all other varieties. For example, if UK English has similar ranks of features as Irish and New Zealand English, then this will produce a high value. But if Swiss English generally has low relationships between feature ranks with other varieties, then this will produce a low value. These uniqueness values are shown in Table 9 for each of the languages with a large number of varieties, calculated using CxG-2 web-based models. Spearman correlations are preferred here instead of Pearson correlations because this reduces the impact of the distance between varieties (which the classifier is trying to maximize).

The uniqueness of each region reflects, at least for non-contiguous languages like English and French, the degree to which a variety belongs in the inner-circle. For example, the top three countries for English are the United States, the UK, and Canada; for French they are France, R\'{e}union (the only French overseas department in the model), and Switzerland. In both cases the uniqueness of varieties with this measure reflects the same scale that categorizations like inner and outer circle are attempting to create. The most unique variety of Spanish is the only non-contiguous variety (from Spain). The interpretation of the rest of the regions on this scale is made more difficult because they are of course densely situated. Notice, also, that while English and French have a scale with higher uniqueness (with starting values of -0.46 and -0.49), both Spanish and Russian have a scale with higher similarity (with ending values of 0.84 and 0.32). Russian has no negative values at all, for example. The most unique varieties of Russian are from Tajikistan, Estonia, and Slovenia. Rather than being inner-circle, as in French and English, these are more peripheral varieties. While this uniqueness measure still reflects an important property of the relationships between varieties, then, its interpretation is complicated by the different behaviour of languages with contiguous or non-contiguous varieties.

The purpose of this section has been to show that the feature weights from the model can also be used to create a general measure of variety uniqueness which reflects an important property of the status of varieties. While qualitative work creates categories like inner-circle or outer-circle, this produces a scale that represents similar intuitions. The difference is that the notion of inner-circle depends on historical and social information about variety areas, with little linguistic analysis, while this scale is entirely linguistic with no historical information whatsoever.

\subsection{Empirical Evidence for World Englishes}

How can we connect data-driven approaches to syntactic variation with qualitative assessments within sociolinguistics? In this section we compare the model of English variation in this paper with traditional classifications from the World Englishes paradigm into inner-circle, outer-circle, and expanding-circle varieties.

First we look at classification accuracy (c.f., Table 5). We expect that inner-circle varieties will be more closely clustered together as they are more closely related and are used in mainly monolingual contexts. There is a significant difference between inner-circle and outer-circle performance in both datasets using a two-tailed t-test ($p=0.0183$ for CC and $p=0.004$ for TW). Upon inspection we see that the outer-circle varieties have higher accuracies, in part because they are more unique.

Second, we look at the degree of fit between the grammar and each regional variety using the relative average frequency: how often do constructions in the grammar occur in each variety? In other words, because the grammar is learned on a different dataset which is likely skewed toward inner-circle varieties, we would expect that the grammar itself would better describe these varieties. A higher average frequency means a better default description (i.e., because the samples are all the same length and so should contain approximately the same number of constructions per sample). We again divide the varieties into inner-circle and outer-circle and test the significance of this difference using a two-tailed t-test: the result is significant ($p=0.0011$ for CC and $p=0.0004$ for TW). In this case, inspection shows that the inner-circle varieties have higher frequencies than the outer-circle varieties.

Third, we look at uniqueness values as calculated in Table 9. First, we see that there is a clear separation between inner-circle and outer-circle varieties, with the exception of South African English. But is the difference significant? Again using a two-tailed t-test there is a significant difference, although to a lesser degree $p=0.024$ for CC).

In all three cases, there is a significant difference between attributes of inner-circle and outer-circle varieties: the proto-typical inner-circle varieties are better described by the grammar but less distinguishable in terms of classification accuracy and in terms of aggregate similarities. There is a consistent and significant distinction, even when the model of varieties of English makes no geographic or sociohistorical assumptions.

\section{Discussion}

This paper has shown (i) that whole-grammar dialectometry and data-driven language mapping can be brought together to produce models capable of predicting the membership of held-out samples with a high degree of accuracy. In addition, we have shown  (ii) that these models do not depend on only a small number of highly predictive variants, (iii) that there is a strong association between classification errors across registers in those languages that are paired across both datasets, (iv) that the models can be used to create reasonable representations of the aggregate similarity between varieties, and (v) that measures of uniqueness based on these models provide an empirical approximation of categorical notions like inner-circle vs. outer-circle varieties. Taken together, these results show that a computational approach to dialectology can overcome the limitations of traditional small-scale methods. The discussion in this section focuses on two questions: First, how do these computational models of dialect relate to previous qualitative understandings of dialect? Second, what does the increased scale and scope of these models mean for interactions between sociolinguistics and computational linguistics?

\subsection{Categorizing Varieties}

At its core, the goal of computational dialectology is to provide precise global-scale models of regional linguistic variation that are both replicable and falsifiable. In other words, these models are \textit{descriptions} of how linguistic structure (specifically, syntax as represented by CxG) varies across national varieties. But we also want to \textit{explain} linguistic variation in historical or social terms: what real-world events caused the spread of these languages in order to create the aggregate relationships that we now observe? While such historical explanations are often \textit{ad hoc}, this paper has attempted to explain synchronic variation using only empirical measures. While it is certainly the case that the concepts used here (predictive accuracy, region similarity, region uniqueness) tell us about varieties, it is not the case that they tell us the same things as traditional qualitative studies. In this case, two clear differences between this paper and traditional approaches to dialectology and dialectometry are (i) the focus on global variation with countries as the smallest spatial unit and (ii) the focus on written as opposed to spoken language.

First, we have a distinction between places (i.e., English used in the United States) and varieties (i.e., American English). There is a claim, whether implicit or explicit, in traditional dialectology that these two are not the same thing. For example, some speakers (older, male, rural, less educated) are taken as more representative than others (younger, urban, immigrant). A farmer born and raised in Kansas is assumed to be a local, a representative of American English; an IT specialist born in India but educated and living in Kansas is not. The argument in this paper, and perhaps in corpus-based research more broadly, is that this starting assumption is problematic. In short, we take American English to be English as used in the United States. We make no effort to exclude certain participants. This approach, then, can be situated within a larger movement away from NORM-based studies \citep{Cheshire2015,Scherrer2016}.

Second, the dialect areas used in this paper ignore distinctions between native speakers and non-native speakers. Similar to the idea of locals vs. non-locals, the claim is that some places that produce a great deal of English data (for example, Nigeria or Malaysia) do not have the same status as American English as sources of ground-truth English data. This distinction is clearly a slippery-slope: while some language learners are not fully fluent, people who use a language like English for regular communicative functions cannot be categorized given \textit{a priori} reasonings. We take this instead as an empirical question: language mapping is used to discover countries where English is regularly and robustly produced and dialect modeling is used to validate that these countries have distinct and predictable varieties. The social status of different English users (i.e., native vs. non-native) is entirely non-empirical and irrelevant. Given that these datasets do not come with individual demographics, however, it is important to also evaluate how well they reflect known demographic properties of the places they are taken to represent in order to ensure the connection between places and syntactic variants \citep{Dunn2019c}.

Third, a distinction is sometimes made between varieties and dialects. For example, outer-circle and expanding-circle dialects are often called varieties. But what is the basis of this distinction? The argument in this paper is simple: the status of Nigerian English or Cameroon French or Angolan Portuguese is an empirical matter. The question is whether we can find these varieties using data-driven language mapping and can model their syntactic profile accurately enough to distinguish them from other varieties consistently across registers.

While previous work in dialectology and dialectometry focuses specifically on variation within individual countries, this paper has focused on global variation across many national varieties. One on the hand, this is important because the seven languages studied in this paper are used around the world: any local study will overlook important interactions. On the other hand, this means that these results are difficult to compare with previous small-scale studies. How could these methods be adapted to traditional problems of, for example, dividing Britain or the United States into dialect regions? First, there is no explicit spatial information provided to the models in this paper because the classes are all pre-defined. On approach would be to use existing sub-national administrative boundaries (such as postal codes) and apply a meta-classifier to evaluate different groupings. Which combinations lead to the highest predictive accuracy? This could be undertaken with the Twitter dataset but not with the web-crawled dataset.

\subsection{Sociolinguistics and Computational Linguistics}

Why should sociolinguistics more broadly care about a computational approach to dialectology? The first reason is simply a matter of descriptive adequacy: the models of variation in this paper have a broad and replicable feature space that is ultimately more meaningful and robust than multivariate models containing only a few features. While the grammars used are not explored further here, quantitative and qualitative evaluations are available elsewhere \citep{d17, d18, d18b, Dunn2019}. These models are more meaningful because they make predictions about categories as a whole (i.e., American English). They are more robust because they are evaluated against held-out samples using predictive accuracy. For both of these reasons, computational models of variation provide more accurate descriptions; this is important for quantitative sociolinguistics, then, simply as an extension of existing methods for discovering externally-conditioned variants (here, conditioned by geography). On the other hand, this approach of combining grammar induction and text classification produces models that, while easily understood in the aggregate, ultimately give us intricate and detailed descriptions that are difficult for human analysts to understand. The question is, do we expect human analysts to have full and complete meta-awareness for all variants in all national varieties of a language?

Beyond this, however, sociolinguistics is currently limited to small-scale studies, as discussed in the introduction. But the languages studied in this paper are used in many countries around the world. Each of these varieties has the potential to influence or be influenced by other distant varieties. In the same way, limiting a study to a handful of constructions ignores most of the functional capability of a language. Thus, current methods provide tiny snapshots of variation. But, moving forward, our ability to further understand syntactic variation and change depends on modeling entire grammars across all relevant varieties. While recent work has increased the number of features in order to produce larger-scale studies \citep{s13, Guy2015}, such features remain language-specific and are defined \textit{a priori}. On the other hand, however, a continued question for work that is bottom-up, such as this paper, is how to evaluate the connection between corpus-based models (which have been shown to be stable, robust, and highly accurate from an internal evaluation) and speech communities in the real world. How can computational descriptions and qualitative fieldwork be better combined?

Given the higher performance of lexical features in this paper, why should work in NLP that is not directly concerned with linguistic variation take a CxG or some other syntactic approach? There is an important distinction between topic variation (i.e., content arising from differences in human geography) and latent variation (i.e., structural variations arising from differences in variety). Any purely-lexical model is unable to distinguish between these two sources of information: Is this text written by someone from New Zealand or is it about New Zealand? Does this Tweet describe a vacation in New Zealand or was it written by a New Zealander on a vacation in the United States? Any model that is unable to distinguish between topical and latent properties within geo-referenced datasets will confuse these two types of cases. On the other hand, this is an incomplete approach the problem: how can we distinguish between topical variation, human geography-based varation, and linguistic variation within lexical items in order to have a better understanding of how these languages are used around the world? This remains a problem for future research.

Why should computational linguistics, and artificial intelligence more broadly, care about dialectology? As computational models become more important to society, it is essential that such models reflect all speakers equally. In spite of this, many models are biased against certain populations: either directly encoding the biases of individuals \citep{Bolukbasi2016} or indirectly encoding a preference for dominant inner-circle varieties \citep{Jurgens2017}. Dialectometry can be used to prevent indirect biases against varieties like Nigerian English or Cameroon French by, first, identifying the relevant varieties that need to be considered and, second, providing a method to optimize language models for region-specific tasks. For example, if we can identify the membership of a sample that is part of an independent text classification problem (i.e., identifying helpful reviews or removing harrassing messages), then we can evaluate the degree to which existing models prefer dominant varieties (i.e., only suggesting reviews written in American English). This is important to ensure that inner-circle dominated training sets do not encode implicit biases against other varieties. It is also important because computational dialectometry can potentially improve equity between varieties in a way that traditional methods cannot.

\section*{Conflict of Interest Statement}
%All financial, commercial or other relationships that might be perceived by the academic community as representing a potential conflict of interest must be disclosed. If no such relationship exists, authors will be asked to confirm the following statement: 

The authors declare that the research was conducted in the absence of any commercial or financial relationships that could be construed as a potential conflict of interest.

\section*{Acknowledgments}
The author would like to thank the editor, Jack Grieve, as well as the anonymous reviewers. This paper has also benefited from audiences at the University of Canterbury, NWAV-46, VarDial-19, LSA 2018 and from on-going conversations with Josh Trampier and Ben Adams.

\section*{Data Availability Statement}

The code and data for this paper can be found in the following locations:

Construction Grammar package: 

\url{https://github.com/jonathandunn/c2xg}

Common Crawl Data collection: 

\url{https://github.com/jonathandunn/common_crawl_corpus}

Language identification code: 

\url{https://github.com/jonathandunn/idNet}

Language identification models:

\url{https://labbcat.canterbury.ac.nz/download/?jonathandunn/idNet_models}

Common Crawl data: 

\url{https://labbcat.canterbury.ac.nz/download/?jonathandunn/CGLU_v3}

Grammar Learning data: 

\url{https://labbcat.canterbury.ac.nz/download/?jonathandunn/CxG_Data_FixedSize}

Experiment code, vectors, and raw results:

\url{https://labbcat.canterbury.ac.nz/download/?jonathandunn/Frontiers_in_AI}

% Please see the availability of data guidelines for more information, at https://www.frontiersin.org/about/author-guidelines#AvailabilityofData

\bibliography{generalizing_dialectometry}

\begin{thebibliography}{70}
\providecommand{\natexlab}[1]{#1}
\expandafter\ifx\csname urlstyle\endcsname\relax
  \providecommand{\doi}[1]{doi:\discretionary{}{}{}#1}\else
  \providecommand{\doi}{doi:\discretionary{}{}{}\begingroup
  \urlstyle{rm}\Url}\fi
\providecommand{\selectlanguage}[1]{\relax}
\providecommand{\bibAnnoteFile}[1]{%
  \IfFileExists{#1}{\begin{quotation}\noindent\textsc{Key:} #1\\
  \textsc{Annotation:}\ \input{#1}\end{quotation}}{}}
\providecommand{\bibAnnote}[2]{%
  \begin{quotation}\noindent\textsc{Key:} #1\\
  \textsc{Annotation:}\ #2\end{quotation}}

\bibitem[{Adams(2015)}]{Adams2015}
Adams, B. (2015).
\newblock {Finding Similar Places using the Observation-to-Generalization Place
  Model}.
\newblock \emph{Journal of Geographical Systems} 17, 137--156
\bibAnnoteFile{Adams2015}

\bibitem[{Adams and McKenzie(2018)}]{Adams2018}
Adams, B. and McKenzie, G. (2018).
\newblock {Crowdsourcing the character of a place: Character-level
  convolutional networks for multilingual geographic text classification}.
\newblock \emph{Transactions in GIS} 22, 394--408
\bibAnnoteFile{Adams2018}

\bibitem[{Argamon and Koppel(2013)}]{ak13}
Argamon, S. and Koppel, M. (2013).
\newblock {A Systemic Functional Approach to Automated Authorship Analysis}.
\newblock \emph{Journal of Law {\&} Policy} 12, 299--315
\bibAnnoteFile{ak13}

\bibitem[{Baroni et~al.(2009)Baroni, Bernardini, Ferraresi, and
  Zanchetta}]{bbfz09}
Baroni, M., Bernardini, S., Ferraresi, A., and Zanchetta, E. (2009).
\newblock {The WaCky Wide Web: A Collection of Very Large Linguistically
  Processed Web-crawled Corpora}.
\newblock \emph{Language Resources and Evaluation} 43, 209--226
\bibAnnoteFile{bbfz09}

\bibitem[{Benko(2014)}]{b14}
Benko, V. (2014).
\newblock {Aranea Yet Another Family of (Comparable) Web Corpora}.
\newblock In \emph{Proceedings of 17th International Conference Text, Speech
  and Dialogue.} (Springer). 257--264
\bibAnnoteFile{b14}

\bibitem[{Bolukbasi et~al.(2016)Bolukbasi, Chang, Zou, Saligrama, and
  Kalai}]{Bolukbasi2016}
Bolukbasi, T., Chang, K.-W., Zou, J., Saligrama, V., and Kalai, A. (2016).
\newblock {Debiasing Word Embedding}.
\newblock \emph{30th Conference on Neural Information Processing Systems} ,
  1--9
\bibAnnoteFile{Bolukbasi2016}

\bibitem[{Calle-Martin and Romero-Barranco(2017)}]{cr17}
Calle-Martin, J. and Romero-Barranco, J. (2017).
\newblock {Third person present tense markers in some varieties of English}.
\newblock \emph{English World-Wide} 38, 77--103
\bibAnnoteFile{cr17}

\bibitem[{Cheshire et~al.(2015)Cheshire, Nortier, and Adger}]{Cheshire2015}
Cheshire, J., Nortier, J., and Adger, D. (2015).
\newblock {Emerging Multiethnolects in Europe}.
\newblock \emph{Queen Mary's Occasional Papers Advancing Linguistics} 33, 1--27
\bibAnnoteFile{Cheshire2015}

\bibitem[{Chomsky(1957)}]{Chomsky1957}
Chomsky, N. (1957).
\newblock \emph{{Syntactic Structures}} (Berlin: Mouton {\&} Co.)
\bibAnnoteFile{Chomsky1957}

\bibitem[{Collins(2012)}]{c12}
Collins, P. (2012).
\newblock {Singular agreement in there existentials: An intervarietal
  corpus-based study}.
\newblock \emph{English World-Wide} 33, 53--68
\bibAnnoteFile{c12}

\bibitem[{Cook and Brinton(2017)}]{Cook2017}
Cook, P. and Brinton, J. (2017).
\newblock {Building and Evaluating Web Corpora Representing National Varieties
  of English}.
\newblock \emph{Language Resources and Evaluation} 51, 643--662
\bibAnnoteFile{Cook2017}

\bibitem[{Davies and Fuchs(2015)}]{df15}
Davies, M. and Fuchs, R. (2015).
\newblock {Expanding horizons in the study of World Englishes with the 1.9
  billion word Global Web-based English Corpus (GloWbE)}.
\newblock \emph{English World-Wide} 36, 1--28
\bibAnnoteFile{df15}

\bibitem[{Donoso and Sanchez(2017)}]{ds17}
Donoso, G. and Sanchez, D. (2017).
\newblock {Dialectometric analysis of language variation in Twitter}.
\newblock \emph{Proceedings of the 4th Workshop on NLP for Similar Languages,
  Varieties and Dialects} 4, 16--25
\bibAnnoteFile{ds17}

\bibitem[{Dunn(2017)}]{d17}
Dunn, J. (2017).
\newblock {Computational Learning of Construction Grammars}.
\newblock \emph{Language {\{}{\&}{\}} Cognition} 9, 254--292
\bibAnnoteFile{d17}

\bibitem[{Dunn(2018{\natexlab{a}})}]{d18b}
Dunn, J. (2018{\natexlab{a}}).
\newblock {Finding Variants for Construction-Based Dialectometry: A
  Corpus-Based Approach to Regional CxGs}.
\newblock \emph{Cognitive Linguistics} 29, 275--311
\bibAnnoteFile{d18b}

\bibitem[{Dunn(2018{\natexlab{b}})}]{d18}
Dunn, J. (2018{\natexlab{b}}).
\newblock {Modeling the Complexity and Descriptive Adequacy of Construction
  Grammars}.
\newblock \emph{In Proceedings of the Society for Computation in Linguistics} ,
  81--90
\bibAnnoteFile{d18}

\bibitem[{Dunn(2019{\natexlab{a}})}]{Dunn2019c}
Dunn, J. (2019{\natexlab{a}}).
\newblock {Frequency vs. Association for Constraint Selection in Usage-Based
  Construction Grammar}.
\newblock In \emph{Proceedings of the Workshop on Cognitive Modeling and
  Computational Linguistics} (Association for Computational Linguistics)
\bibAnnoteFile{Dunn2019c}

\bibitem[{Dunn(2019{\natexlab{b}})}]{Dunn2019a}
Dunn, J. (2019{\natexlab{b}}).
\newblock {Modeling Global Syntactic Variation in English Using Dialect
  Classification}.
\newblock In \emph{Proceedings of NAACL 2019 Sixth Workshop on NLP for Similar
  Languages, Varieties and Dialects} (Association for Computational
  Linguistics), 42--53
\bibAnnoteFile{Dunn2019a}

\bibitem[{Dunn and Adams(2019)}]{Dunn2019}
Dunn, J. and Adams, B. (2019).
\newblock {Mapping Languages and Demographics with Georeferenced Corpora}.
\newblock In \emph{Proceedings of Geocomputation 2019}. 16 pages.
\bibAnnoteFile{Dunn2019}

\bibitem[{Dunn et~al.(2016)Dunn, Argamon, Rasooli, and Kumar}]{dark16}
Dunn, J., Argamon, S., Rasooli, A., and Kumar, G. (2016).
\newblock {Profile-based authorship analysis}.
\newblock \emph{Literary and Linguistic Computing} 31, 689--710
\bibAnnoteFile{dark16}

\bibitem[{Ginter et~al.(2017)Ginter, Hajiˆc, and Luotolahti}]{ghl17}
Ginter, F., Hajiˆc, J., and Luotolahti, J. (2017).
\newblock \emph{{CoNLL 2017 Shared Task - Automatically Annotated Raw Texts and
  Word Embeddings}} (Faculty of Mathematics and Physics, Charles University:
  LINDATCLARIN digital library at the Institute of Formal and Applied
  Linguistics (FAL))
\bibAnnoteFile{ghl17}

\bibitem[{Goldberg(2006)}]{g06a}
Goldberg, A. (2006).
\newblock \emph{{Constructions at work: The nature of generalization in
  language}} (Oxford: Oxford University Press)
\bibAnnoteFile{g06a}

\bibitem[{Goldhahn et~al.(2012)Goldhahn, Eckart, and Quasthoff}]{geq12}
Goldhahn, D., Eckart, T., and Quasthoff, U. (2012).
\newblock {Building Large Monolingual Dictionaries at the Leipzig Corpora
  Collection From 100 to 200 Languages}.
\newblock In \emph{Proceedings of the Eighth Conference on Language Resources
  and Evaluation} (European Language Resources Association), 759--765
\bibAnnoteFile{geq12}

\bibitem[{Goldsmith(2015)}]{Goldsmith2015}
Goldsmith, J. (2015).
\newblock {Towards a new empiricism for linguistics}.
\newblock In \emph{Empiricism and Language Learnability}, eds. N.~Chater,
  A.~Clark, J.~Goldsmith, and A.~Perfors (Oxford: Oxford University Press).
  58--105
\bibAnnoteFile{Goldsmith2015}

\bibitem[{Grafmiller and Szmrecsanyi(2018)}]{gs18}
Grafmiller, J. and Szmrecsanyi, B. (2018).
\newblock {Mapping out particle placement in Englishes around the world A study
  in comparative sociolinguistic analysis}.
\newblock \emph{Language Variation and Change} 30, 385--412
\bibAnnoteFile{gs18}

\bibitem[{Grieve(2011)}]{Grieve2011}
Grieve, J. (2011).
\newblock {A regional analysis of contraction rate in written Standard American
  English}.
\newblock \emph{International Journal of Corpus Linguistics} 16, 514--546
\bibAnnoteFile{Grieve2011}

\bibitem[{Grieve(2012)}]{Grieve2012}
Grieve, J. (2012).
\newblock {A statistical analysis of regional variation in adverb position in a
  corpus of written Standard American English}.
\newblock \emph{Corpus Linguistics and Linguistic Theory} 8, 39--72
\bibAnnoteFile{Grieve2012}

\bibitem[{Grieve(2013)}]{g13}
Grieve, J. (2013).
\newblock {A statistical comparison of regional phonetic and lexical variation
  in American English}.
\newblock \emph{Literary and Linguistic Computing} 28, 82--107
\bibAnnoteFile{g13}

\bibitem[{Grieve(2016)}]{g16}
Grieve, J. (2016).
\newblock \emph{{Regional variation in written American English}} (Cambridge,
  UK: Cambridge University Press)
\bibAnnoteFile{g16}

\bibitem[{Grieve et~al.(2011)Grieve, Speelman, and Geeraerts}]{gsg11}
Grieve, J., Speelman, D., and Geeraerts, D. (2011).
\newblock {A statistical method for the identification and aggregation of
  regional linguistic variation}.
\newblock \emph{Language Variation {\&} Change} 23, 1--29
\bibAnnoteFile{gsg11}

\bibitem[{Grieve et~al.(2013)Grieve, Speelman, and Geeraerts}]{JackGrieve2013}
Grieve, J., Speelman, D., and Geeraerts, D. (2013).
\newblock {A multivariate spatial analysis of vowel formants in American
  English}.
\newblock \emph{Journal of Linguistic Geography} 1, 31--51
\bibAnnoteFile{JackGrieve2013}

\bibitem[{Guy and Oushiro(2015)}]{Guy2015}
Guy, G. and Oushiro, L. (2015).
\newblock {The effect of salience on co-variation in Brazilian Portuguese.}
\newblock \emph{University of Pennsylvania Working Papers in Linguistics} 21,
  Paper 18
\bibAnnoteFile{Guy2015}

\bibitem[{Heeringa(2004)}]{h04}
Heeringa, W. (2004).
\newblock \emph{{Measuring dialect pronunciation differences using Levenshtein
  distance}} (Groningen, Netherlands)
\bibAnnoteFile{h04}

\bibitem[{Hirst and Feiguina(2007)}]{hf07}
Hirst, G. and Feiguina, O. (2007).
\newblock {Bigrams of syntactic labels for authorship discrimination of short
  texts}.
\newblock \emph{Literary and Linguistic Computing} 22, 405--417
\bibAnnoteFile{hf07}

\bibitem[{Hulden et~al.(2015)Hulden, Silfverberg, and Francom}]{hsf15}
Hulden, M., Silfverberg, M., and Francom, J. (2015).
\newblock {Kernel Density Estimation for Text-Based Geolocation}.
\newblock In \emph{Proceedings of the Twenty-Ninth AAAI Conference on
  Artificial Intelligence} (Association for the Advancement of Artificial
  Intelligence), 145--150
\bibAnnoteFile{hsf15}

\bibitem[{Joachims(1998)}]{j98}
Joachims, T. (1998).
\newblock {Text categorization with support vector machines: Learning with many
  relevant features}.
\newblock In \emph{Proceedings of the European Conference on Machine Learning}
  (Berlin: Springer). 137--142
\bibAnnoteFile{j98}

\bibitem[{Jurgens et~al.(2017)Jurgens, Tsvetkov, Jurafsky, and
  Jurafsy}]{Jurgens2017}
Jurgens, D., Tsvetkov, Y., Jurafsky, D., and Jurafsy, D. (2017).
\newblock {Incorporating Dialectal Variability for Socially Equitable Language
  Identification}.
\newblock In \emph{Proceedings of the Annual Meeting of the Association for
  Computational Linguistics} (Association for Computational Linguistics),
  51--57
\bibAnnoteFile{Jurgens2017}

\bibitem[{Kachru(1990)}]{k90}
Kachru, B. (1990).
\newblock \emph{{The Alchemy of English The spread, functions, and models of
  non-native Englishes}} (Urbana-Champaign: University of Illinois Press)
\bibAnnoteFile{k90}

\bibitem[{Kachru(1982)}]{Kachru1982}
Kachru, B.~e. (1982).
\newblock \emph{{The Other tongue: English across cultures.}}
  (Urbana-Champaign: University of Illinois Press)
\bibAnnoteFile{Kachru1982}

\bibitem[{Koppel et~al.(2007)Koppel, Schler, and Bonchek-Dokow}]{ksb07}
Koppel, M., Schler, J., and Bonchek-Dokow, E. (2007).
\newblock {Measuring differentiability: Unmasking pseudonymous authors}.
\newblock \emph{Journal of Machine Learning Research} 8, 1261--1276
\bibAnnoteFile{ksb07}

\bibitem[{Kretzschmar(1992)}]{k92}
Kretzschmar, W.~A. (1992).
\newblock {Isoglosses and predictive modeling}.
\newblock \emph{American Speech} 67, 227--249
\bibAnnoteFile{k92}

\bibitem[{Kretzschmar(1996)}]{k96}
Kretzschmar, W.~A. (1996).
\newblock {Quantitative areal analysis of dialect features}.
\newblock \emph{Language Variation {\&} Change} 8, 13--39
\bibAnnoteFile{k96}

\bibitem[{Kretzschmar et~al.(2014)Kretzschmar, Juuso, and Bailey}]{kjb14}
Kretzschmar, W.~A., Juuso, I., and Bailey, C. (2014).
\newblock {Computer simulation of dialect feature diffusion}.
\newblock \emph{Journal of Linguistic Geography} 2, 41--57
\bibAnnoteFile{kjb14}

\bibitem[{Kroon et~al.(2018)Kroon, Medvedeva, and Plank}]{k}
Kroon, M., Medvedeva, M., and Plank, B. (2018).
\newblock {When Simple n-gram Models Outperform Syntactic Approaches
  Discriminating between Dutch and Flemish}.
\newblock \emph{Proceedings of the Fifth Workshop on NLP for Similar Languages,
  Varieties and Dialects} , 225--244
\bibAnnoteFile{k}

\bibitem[{Kruger and van Rooy(2018)}]{kv18}
Kruger, H. and van Rooy, B. (2018).
\newblock {Register variation in written contact varieties of English: A
  multidimensional analysis}.
\newblock \emph{English World-Wide} 39, 214--242
\bibAnnoteFile{kv18}

\bibitem[{Labov et~al.(2005)Labov, Ash, and Boberg}]{lab05}
Labov, W., Ash, S., and Boberg, C. (2005).
\newblock \emph{{The Atlas of North American English: Phonetics, phonology and
  sound change}} (Berlin: De Gruyter Mouton)
\bibAnnoteFile{lab05}

\bibitem[{Langacker(2008)}]{l08}
Langacker, R. (2008).
\newblock \emph{{Cognitive Grammar A basic introduction}} (Oxford: Oxford
  University Press)
\bibAnnoteFile{l08}

\bibitem[{Lourentzou et~al.(2017)Lourentzou, Morales, and Zhai}]{lmz17}
Lourentzou, I., Morales, A., and Zhai, C. (2017).
\newblock {Textbased geolocation prediction of social media users with neural
  networks}.
\newblock In \emph{Proceedings of 2017 IEEE International Conference on Big
  Data} (Institute of Electrical and Electronics Engineers), 696--705
\bibAnnoteFile{lmz17}

\bibitem[{Majliˇs and ZdenˇekˇZabokrtsky(2012)}]{m12}
Majliˇs, M. and ZdenˇekˇZabokrtsky (2012).
\newblock {Language Richness of the Web}.
\newblock In \emph{Proceedings of the International Conference on Language
  Resources and Evaluation} (European Language Resources Association),
  2927--2934
\bibAnnoteFile{m12}

\bibitem[{Nerbonne(2006)}]{n06}
Nerbonne, J. (2006).
\newblock {Identifying linguistic structure in aggregate comparison}.
\newblock \emph{Literary and Linguistic Computing} 21, 463--476
\bibAnnoteFile{n06}

\bibitem[{Nerbonne(2009)}]{n09}
Nerbonne, J. (2009).
\newblock {Data-driven dialectology}.
\newblock \emph{Language and Linguistics Compass} 3, 175--198
\bibAnnoteFile{n09}

\bibitem[{Nerbonne and Kretzschmar(2013)}]{nk13}
Nerbonne, J. and Kretzschmar, W. (2013).
\newblock {Dialectometry+ +}.
\newblock \emph{Literary and Linguistic Computing} 28, 2--12
\bibAnnoteFile{nk13}

\bibitem[{Rangel et~al.(2017)Rangel, Rosso, Potthast, and Stein}]{Rangel2017}
Rangel, F., Rosso, P., Potthast, M., and Stein, B. (2017).
\newblock {Overview of the 5th Author Profiling Task at PAN 2017: Gender and
  Language Variety Identification in Twitter}.
\newblock In \emph{CLEF 2017 Labs and Workshops, Notebook Papers. CEUR Workshop
  Proceedings} (CEUR-WS.org), vol. 1866
\bibAnnoteFile{Rangel2017}

\bibitem[{Ruette and Speelman(2014)}]{Ruette2014}
Ruette, T. and Speelman, D. (2014).
\newblock {Transparent aggregation of variables with Individual Differences
  Scaling}.
\newblock \emph{Literary and Linguistic Computing} 29, 89--106
\bibAnnoteFile{Ruette2014}

\bibitem[{Sanders(2007)}]{s07}
Sanders, N.~C. (2007).
\newblock {Measuring syntactic difference in British English}.
\newblock \emph{Proceedings of the ACL 2007 Student Research Workshop} 45, 1--6
\bibAnnoteFile{s07}

\bibitem[{Sanders(2010)}]{s10}
Sanders, N.~C. (2010).
\newblock \emph{{A statistical method for syntactic dialectometry}}
  (Bloomington: dissertation)
\bibAnnoteFile{s10}

\bibitem[{Scherrer and Stoeckle(2016)}]{Scherrer2016}
Scherrer, Y. and Stoeckle, P. (2016).
\newblock {A quantitative approach to Swiss German - Dialectometric analyses
  and comparison of linguistic levels.}
\newblock \emph{Dialectologia et Geolinguistica} 24, 92--125
\bibAnnoteFile{Scherrer2016}

\bibitem[{Schilk and Schaub(2016)}]{ss16}
Schilk, M. and Schaub, S. (2016).
\newblock {Noun phrase complexity across varieties of English: Focus on
  syntactic function and text type}.
\newblock \emph{English World-Wide} 37, 58--85
\bibAnnoteFile{ss16}

\bibitem[{Skadiņ{\v{s}} et~al.(2014)Skadiņ{\v{s}}, Tiedemann, Rozis, and
  Deksne}]{Skadins2014}
Skadiņ{\v{s}}, R., Tiedemann, J., Rozis, R., and Deksne, D. (2014).
\newblock {Billions of Parallel Words for Free}.
\newblock In \emph{Proceedings of the International Conference on Language
  Resources and Evaluation} (European Language Resources Association)
\bibAnnoteFile{Skadins2014}

\bibitem[{Szmrecsanyi(2009)}]{s09}
Szmrecsanyi, B. (2009).
\newblock {Corpus-based dialectometry Aggregate morphosyntactic variability in
  British English dialects}.
\newblock \emph{International Journal of Humanities and Arts Computing} 2,
  279--296
\bibAnnoteFile{s09}

\bibitem[{Szmrecsanyi(2013)}]{s13}
Szmrecsanyi, B. (2013).
\newblock \emph{{Grammatical variation in British English dialects A study in
  corpus-based dialectometry}} (Cambridge: Cambridge University Press)
\bibAnnoteFile{s13}

\bibitem[{Szmrecsanyi(2014)}]{s14}
Szmrecsanyi, B. (2014).
\newblock {Forests, trees, corpora, and dialect grammars}.
\newblock In \emph{Aggregating dialectology, typology, and register analysis
  Linguistic variation in text and speech.}, eds. A.~dialectology ), Typology,
  register analysis Linguistic variation~in text, and B.~M. D.~G. Speech
  (Berlin: Mouton De Gruyter). 89--112
\bibAnnoteFile{s14}

\bibitem[{Szmrecsanyi et~al.(2016)Szmrecsanyi, Grafmiller, Heller, and
  Rothlisberger}]{sghr16}
Szmrecsanyi, B., Grafmiller, J., Heller, B., and Rothlisberger, M. (2016).
\newblock {Around the world in three alternations Modeling syntactic variation
  in varieties of English}.
\newblock \emph{English World-Wide} 37, 109--137
\bibAnnoteFile{sghr16}

\bibitem[{Tamaredo(2018)}]{t18}
Tamaredo, I. (2018).
\newblock {Pronoun omission in high-contact varieties of English Complexity
  versus efficiency}.
\newblock \emph{English World-Wide} 39, 85--110
\bibAnnoteFile{t18}

\bibitem[{Tiedemann(2012)}]{Tiedemann2012}
Tiedemann, J. (2012).
\newblock {Parallel Data, Tools and Interfaces in OPUS}.
\newblock In \emph{Proceedings of the International Conference on Language
  Resources and Evaluation} (European Language Resources Association)
\bibAnnoteFile{Tiedemann2012}

\bibitem[{{United Nations}(2017)}]{UnitedNations2017}
{United Nations} (2017).
\newblock \emph{{World Population Prospects: The 2017 Revision, DVD Edition}}
  (United Nations Population Division)
\bibAnnoteFile{UnitedNations2017}

\bibitem[{Wieling and Nerbonne(2011)}]{wn11}
Wieling, M. and Nerbonne, J. (2011).
\newblock {Bipartite spectral graph partitioning for clustering dialect
  varieties and detecting their linguistic features}.
\newblock \emph{Computer Speech {\&} Language} 25, 700--715
\bibAnnoteFile{wn11}

\bibitem[{Wieling and Nerbonne(2015)}]{wn15}
Wieling, M. and Nerbonne, J. (2015).
\newblock {Advances in dialectometry}.
\newblock \emph{Annual Review of Linguistics} 1, 243--264
\bibAnnoteFile{wn15}

\bibitem[{Wing and Baldridge(2014)}]{Wing2014}
Wing, B. and Baldridge, J. (2014).
\newblock {Hierarchical Discriminative Classification for Text-Based
  Geolocation}.
\newblock In \emph{Proceedings of the Conference on Empirical Methods in NLP}
  (Association for Computational Linguistics), 336--348
\bibAnnoteFile{Wing2014}

\bibitem[{Zenner et~al.(2012)Zenner, Speelman, and Geeraerts}]{Zenner2012}
Zenner, E., Speelman, D., and Geeraerts, D. (2012).
\newblock {Cognitive Sociolinguistics meets loanword research: Measuring
  variation in the success of anglicisms in Dutch}.
\newblock \emph{Cognitive Linguistics} 23, 749--792.
\newblock \doi{10.1515/cog-2012-0023}
\bibAnnoteFile{Zenner2012}

\end{thebibliography}
\bibliographystyle{frontiersinSCNS_ENG_HUMS} % for Science, Engineering and Humanities and Social Sciences articles, for Humanities and Social Sciences articles please include page numbers in the in-text citations
%\bibliographystyle{frontiersinHLTH&FPHY} % for Health, Physics and Mathematics articles

%%% Make sure to upload the bib file along with the tex file and PDF
%%% Please see the test.bib file for some examples of references

\section*{Figure captions}

%%% Please be aware that for original research articles we only permit a combined number of 15 figures and tables, one figure with multiple subfigures will count as only one figure.
%%% Use this if adding the figures directly in the mansucript, if so, please remember to also upload the files when submitting your article
%%% There is no need for adding the file termination, as long as you indicate where the file is saved. In the examples below the files (logo1.eps and logos.eps) are in the Frontiers LaTeX folder
%%% If using *.tif files convert them to .jpg or .png
%%%  NB logo1.eps is required in the path in order to correctly compile front page header %%%

%%% If you are submitting a figure with subfigures please combine these into one image file with part labels integrated.
%%% If you don't add the figures in the LaTeX files, please upload them when submitting the article.
%%% Frontiers will add the figures at the end of the provisional pdf automatically
%%% The use of LaTeX coding to draw Diagrams/Figures/Structures should be avoided. They should be external callouts including graphics.

\begin{table}
\begin{center}
	\def\arraystretch{1.5}
	\begin{tabular}{|cl|cl|cl|}
		\hline
AR & Argentina & HN & Honduras & PT & Portugal \\
AU & Australia & HT & Haiti & PW & Palau \\
AZ & Azerbaijan & IE & Ireland & PY & Paraguay \\
BE & Belgium & IN & India & QA & Qatar \\
BF & Burkina Faso & IQ & Iraq & RE & R\'{e}union \\
BG & Bulgaria & JO & Jordan & RU & Russia \\
BR & Brazil & KG & Kyrgyzstan & SI & Slovenia \\
BY & Belarus & KW & Kuwait & SN & Senegal \\
CA & Canada & KZ & Kazakhstan & SO & Somalia \\
CH & Switzerland & LT & Lithuania & SV & El Salvador \\
CL & Chile & LU & Luxembourg & SY & Syria \\
CM & Cameroon & LV & Latvia & TJ & Tajikistan \\
CO & Colombia & MD & Moldova & TN & Tunisia \\
CR & Costa Rica & MX & Mexico & UA & Ukraine \\
CU & Cuba & MY & Malaysia & UK & United Kingdom \\
CV & Cabo Verde & NC & New Caledonia & US & USA \\
DZ & Algeria & NG & Nigeria & UY & Uruguay \\
EC & Ecuador & NI & Nicaragua & UZ & Uzbekistan \\
EE & Estonia & NZ & New Zealand & VE & Venezuela \\
EG & Egypt & PA & Panama & ZA & South Africa \\
ES & Spain & PE & Peru & ~ & ~ \\
FR & France & PF & French Polynesia & ~ & ~ \\
GD & Grenada & PH & Philippines & ~ & ~ \\
GE & Georgia & PK & Pakistan & ~ & ~ \\
GT & Guatemala & PS & Palestine & ~ & ~ \\
		\hline
	\end{tabular}
	\caption{Appendix A: Abbreviations for Countries}
	\label{tab:1}
	\end{center}
\end{table}

\begin{table}
\begin{center}
	\def\arraystretch{1.5}
	\begin{tabular}{|c|ccc|c|ccc|}
		\hline
\textbf{Arabic} & \textbf{Prec (CC)} & \textbf{Recall (CC)} & \textbf{F1 (CC)} & ~ & \textbf{Prec (TW)} & \textbf{Recall (TW)} & \textbf{F1 (TW)} \\
\hline
AE & 1.00 & 1.00 & 1.00 & AE & -- & -- & -- \\
DZ & -- & -- & -- & DZ & 0.98 & 0.98 & 0.98 \\
EG & -- & -- & -- & EG & 0.99 & 0.99 & 0.99 \\
IQ & -- & -- & -- & IQ & 0.97 & 0.97 & 0.97 \\
JO & -- & -- & -- & JO & 0.97 & 0.97 & 0.97 \\
KW & -- & -- & -- & KW & 0.94 & 0.96 & 0.95 \\
PS & 1.00 & 1.00 & 1.00 & PS & -- & -- & -- \\
QA & 1.00 & 0.99 & 1.00 & QA & -- & -- & -- \\
RU & -- & -- & -- & RU & 1.00 & 1.00 & 1.00 \\
SY & 1.00 & 0.99 & 0.99 & SY & 0.97 & 0.97 & 0.97 \\
		\hline
\textbf{German} & \textbf{Prec (CC)} & \textbf{Recall (CC)} & \textbf{F1 (CC)} & ~ & \textbf{Prec (TW)} & \textbf{Recall (TW)} & \textbf{F1 (TW)} \\
\hline
AT & 0.95 & 0.95 & 0.95 & AT & 0.96 & 0.91 & 0.93 \\
CH & 0.94 & 0.94 & 0.94 & CH & -- & -- & -- \\
DE & 0.92 & 0.91 & 0.92 & DE & 0.95 & 0.98 & 0.97 \\
LU & 0.97 & 0.96 & 0.96 & LU & -- & -- & -- \\
PL & 0.98 & 0.98 & 0.98 & PL & -- & -- & -- \\
\hline
\textbf{Portuguese} & \textbf{Prec (CC)} & \textbf{Recall (CC)} & \textbf{F1 (CC)} & ~ & \textbf{Prec (TW)} & \textbf{Recall (TW)} & \textbf{F1 (TW)} \\
\hline
AO & 0.99 & 0.99 & 0.99 & AO & -- & -- & -- \\
BR & 1.00 & 1.00 & 1.00 & BR & 1.00 & 1.00 & 1.00 \\
CV & 0.98 & 0.97 & 0.98 & CV & -- & -- & -- \\
PT & 0.99 & 0.99 & 0.99 & PT & 1.00 & 0.99 & 1.00 \\
		\hline
	\end{tabular}
	\caption{Appendix 2: Classification Performance for Arabic (Top), German (Middle), and Portuguese (Bottom), Web and Twitter Corpora with CxG-2 Features}
	\label{tab:1}
	\end{center}
\end{table}

\end{document}